\documentclass[]{assets/matterlab}
\usepackage{microtype}
\usepackage{graphicx}
\usepackage{booktabs}
\usepackage{float}
\usepackage{placeins}
\usepackage{needspace}
\usepackage{xurl}
\usepackage{hyperref}
\usepackage{titlesec}

\usepackage{amsmath}
\usepackage{amssymb}
\usepackage{mathtools}
\usepackage{amsthm}
\usepackage{bm}
\usepackage{algorithm}
\usepackage{algpseudocode}
\usepackage[noabbrev,nameinlink]{cleveref}

\crefname{figure}{Fig.}{Figs.}
\Crefname{figure}{Fig.}{Figs.}
\crefname{table}{Table}{Tables}
\Crefname{table}{Table}{Tables}
\usepackage{amsmath,amsfonts,bm}

\def\eqref#1{equation~\ref{#1}}
\def\1{\bm{1}}

\DeclareMathAlphabet{\mathsfit}{\encodingdefault}{\sfdefault}{m}{sl}
\SetMathAlphabet{\mathsfit}{bold}{\encodingdefault}{\sfdefault}{bx}{n}

\usepackage{orcidlink}
\usepackage{comment}
\usepackage[version=4]{mhchem}
\usepackage{longtable}
\usepackage{array}
\renewcommand{\arraystretch}{1.7}
\usepackage{multirow}           % tabular cells spanning multiple rows
\usepackage{amsfonts}           % blackboard math symbols
\usepackage{nicefrac}
\usepackage{duckuments}         % sample images
\usepackage{thmtools,thm-restate}
\usepackage{enumitem}
\usepackage{acro}
\acsetup{first-style=long-short}
\DeclareAcronym{oled}{short=OLED,long={organic light-emitting diode}}
\DeclareAcronym{tadf}{short=TADF,long={thermally activated delayed fluorescence}}
\DeclareAcronym{gpt2}{short=GPT-2,long={generative pre-trained transformer 2}}
\DeclareAcronym{homo-lumo}{short=HOMO--LUMO,long={highest occupied molecular orbital--lowest unoccupied molecular orbital}}
\DeclareAcronym{tddft}{short=TD-DFT,long={time-dependent density functional theory}}
\DeclareAcronym{dft}{short=DFT,long={density functional theory}}
\DeclareAcronym{smiles}{short=SMILES,long={simplified molecular-input line-entry system}}
\DeclareAcronym{ir}{short=IR,long={infrared}}
\DeclareAcronym{uv}{short=UV,long={ultraviolet}}
\DeclareAcronym{dcm}{short=DCM,long={dichloromethane}}
\DeclareAcronym{llm4sd}{short=LLM4SD,long={Large Language Models for Scientific Discovery}}
\DeclareAcronym{crest}{short=CREST,long={Conformer--Rotamer Ensemble Sampling Tool}}
\DeclareAcronym{gfn2xtb}{short=GFN2-xTB,long={second-generation geometry, frequency, noncovalent, extended tight-binding}}
\DeclareAcronym{rgbsa}{short=rGBSA,long={rigid generalized Born surface area}}
\DeclareAcronym{rijcosx}{short=RIJCOSX,long={resolution-of-identity approximation and chain-of-spheres exchange}}
\DeclareAcronym{cpcm}{short=CPCM,long={conductor-like polarizable continuum model}}
\DeclareAcronym{arm}{short=ARM,long={autoregressive language modelling}}
\DeclareAcronym{mlm}{short=MLM,long={masked language modelling}}
\DeclareAcronym{plm}{short=PLM,long={permutation language modelling}}
\DeclareAcronym{sa}{short=SA,long={synthetic accessibility}}
\usepackage{xcolor,colortbl}
\usepackage{tikz}
\usepackage{tikz-cd}
\usepackage{caption}
\usepackage{subcaption}
\usepackage{listings}
\usepackage{gensymb}
\usepackage{textcomp} % Alternative to gensymb
\DeclareUnicodeCharacter{2212}{\ensuremath{-}}

\newcommand{\addressCHEM}{Department of Chemistry, University of Toronto, Lash Miller Chemical Laboratories, 80 St. George Street, ON M5S 3H6, Toronto, Canada}
\newcommand{\addressAC}{Acceleration Consortium, 700 University Ave., M7A 2S4, Toronto, Canada}
\newcommand{\addressCS}{Department of Computer Science, University of Toronto, 40 St George St., Toronto, ON M5S 2E4, Canada}
\newcommand{\addressVECTOR}{Vector Institute for Artificial Intelligence, W1140-108 College St., Schwartz Reisman Innovation
Campus, Toronto, ON M5G 0C6, Canada}
\newcommand{\addressMSE}{Department of Materials Science \& Engineering, University of Toronto, 184 College St., M5S 3E4, Toronto, Canada}
\newcommand{\addressCHEMENG}{Department of Chemical Engineering \& Applied Chemistry, University of Toronto, 200 College St. ON M5S 3E5, Toronto, Canada}
\newcommand{\addressCIFAR}{Senior Fellow, Canadian Institute for Advanced Research (CIFAR), 661 University Ave., M5G 1M1, Toronto, Canada}
\newcommand{\addressNVIDIA}{NVIDIA, 431 King St W \#6th, M5V 1K4, Toronto, Canada}

\newcommand{\acknowAC}{This research is part of the University of Toronto’s Acceleration Consortium, which receives funding from the CFREF-2022-00042 Canada First Research Excellence Fund.}

\newcommand{\acknowSciNet}[1][Trillium supercomputer]{Computations were performed on the #1 at the SciNet HPC Consortium.  SciNet is funded by Innovation, Science and Economic Development Canada; the Digital Research Alliance of Canada; the Ontario Research Fund: Research Excellence; and the University of Toronto.}

\newcommand{\acknowCalcQueb}{Computations were made on the supercomputer Narval from École de technologie supérieure, managed by Calcul Québec and the Digital Research Alliance of Canada. The operation of this supercomputer is funded by the Canada Foundation for Innovation (CFI), Ministère de l’Économie, des Sciences et de l’Innovation du Québec (MESI) and le Fonds de recherche du Québec – Nature et technologies (FRQ-NT).}

\title{De novo molecular generation with optical property preconditioning at the token level}

\author[1,2,\dagger]{Hazohe Huang}
\author[3, 2,\dagger, \ddagger]{Manuel Gonzalez Lastre}

\author[4]{Hyun Suk Park}
\author[4,2]{Jorge A. Campos-Gonzalez-Angulo}
\author[1]{Xinjian Liu}
\author[4,1,5,6,7,2,8,9,*]{Al\'an Aspuru-Guzik}

\affiliation[1]{\addressCS}
\affiliation[2]{\addressVECTOR}

\affiliation[3]{Departamento de F\'isica Te\'orica de la Materia Condensada, Universidad Aut\'onoma de Madrid, 28049, Madrid, Spain.}

\affiliation[4]{\addressCHEM}

\affiliation[5]{\addressMSE}
\affiliation[6]{\addressCHEMENG}
\affiliation[7]{\addressAC}

\affiliation[8]{\addressCIFAR}
\affiliation[9]{\addressNVIDIA}

\contribution[\dagger]{These authors contributed equally}
\contribution[\ddagger]{Work done during an internship at Vector Institute.}

\abstract{
Designing \ac{oled} molecules with targeted optical properties remains challenging due to the scarcity of high-quality data and the limited reliability of conditional control in generative models across chemical motifs. Here, we benchmark a token-conditioned autoregressive language model for OLED molecular generation in a realistic low-data regime. A \ac{gpt2} model is pretrained on large chemical corpora, augmented with discrete property tokens, and fine-tuned using multi-task optimisation. Conditioning targets vertical absorption energy and oscillator strength, with the \ac{homo-lumo} gap included as an auxiliary electronic descriptor.
Generated molecules are evaluated at the \ac{tddft} level to assess distributional fidelity and controllability. The generated library reproduces the dominant optical-property support of the training distribution while shifting towards lower molecular weight and fewer heavy atoms. Token-level control is consistently directional across conditioning bins, but is not fully orthogonal and exhibits local calibration irregularities. A chemotype-resolved analysis further shows that controllability depends strongly on local electronic environments: moderately conjugated aromatic-carbon motifs are associated with improved joint target satisfaction, whereas electron-withdrawing motifs, particularly aryl nitriles, show systematic red-shifting and reduced controllability.
These results establish a quantitative benchmark for conditional OLED molecular generation and show that model reliability must be assessed in chemically meaningful subspaces rather than from aggregate property distributions alone.
}

\date{\today}
\correspondence{Al\'an Aspuru-Guzik at \email{alan@aspuru.com}}

\metadata[Code]{
\url{https://github.com/aspuru-guzik-group/oled_generation_optical_preconditioning}
}

\begin{document}

\maketitle

%%%

% Here we use \input so that it continues on the same page without a page break
% \include inserts a forced page break before and after the included file
\section{Introduction}

\Acp{oled} are central to modern optoelectronic technologies because their optical response can be tuned directly through molecular structure.\cite{gomez-bombarelliDesignEfficientMolecular2016} Advances in organic fluorescent, phosphorescent, and \ac{tadf} emitters have shown that controlling conjugation, donor--acceptor architectures, and excited-state energetics enables efficient light emission across a broad spectral range.\cite{Itoh_2012,gomez-bombarelliDesignEfficientMolecular2016} Beyond display technologies, structurally tunable organic chromophores are also important in broader optoelectronic and photonic settings, including optical sensing, imaging, and light-harvesting applications, where spectral control and synthetic flexibility are key design advantages.\cite{Greenman2022}

OLED discovery is increasingly driven by data-enabled screening and design. Large-scale virtual-screening studies have shown that combinatorial molecular libraries can be searched effectively when coupled to quantum chemistry and machine learning.\cite{gomez-bombarelliDesignEfficientMolecular2016,Greenman2022} More broadly, optical informatics has been dominated by property-prediction models based on molecular fingerprints, atomistic descriptors, convolutional architectures, and graph neural networks.\cite{jungAutomaticPredictionPeak2024,heidChempropMachineLearning2024,souzaPredictingFluorescenceEmission2025} These methods accelerate the estimation of observables such as absorption energies, emission wavelengths, and oscillator strengths. However, they do not, by themselves, solve the inverse design problem of generating new molecules with targeted optical properties. In practice, inverse design in this domain remains constrained by the scarcity of high-quality optical data and by inconsistent controllability across chemical subspaces. We therefore study conditional generation in a realistic low-data regime, focusing on targets that are both device-relevant and computationally tractable for large-scale screening.

Generative modelling offers a more direct route to inverse design, but several challenges remain. Earlier work explored latent-space optimization with variational autoencoders and related representation-learning strategies, in which property predictors guide the search within a learned molecular embedding.\cite{altae-tranLowDataDrug2017,gomez2018automatic} Subsequent conditional and guided-generation approaches extended this idea by steering generation toward desired objectives through property prompts, fragment constraints, or target-specific conditions.\cite{yangCMGNConditionalMolecular2023,bornRegressionTransformerEnables2023} Transformer-based molecular language models have further expanded this space. Models such as MolGPT,\cite{molgpt} LlaMol,\cite{zhangLlaMolDynamicMulticonditional2024} Token-Mol,\cite{wangTokenMolTargetguidedMolecular2025} and the Regression Transformer show that sequence-based architectures can support scaffold conditioning, multi-property prompting, and joint predictive--generative learning.\cite{bornRegressionTransformerEnables2023} Recent work has also shown that autoregressive generative modelling remains competitive beyond string-based representations: Quetzal enables scalable atom-by-atom 3D molecule generation and achieves quality comparable to diffusion-based 3D generators, while naturally supporting variable-size tasks such as scaffold completion.\cite{chengScalableAutoregressive3DMolecule2025} In parallel, frameworks such as \ac{llm4sd} have demonstrated value for extracting interpretable structure--property rules, but remain focused primarily on molecular property prediction rather than de novo generation.\cite{zhengLargeLanguageModels2025}

Within the optical and OLED domain, generative design remains less mature than predictive modelling. Recent studies have begun to address this gap, including OLED-oriented workflows such as LumiGen.\cite{niuDataDrivenOLEDCandidate2025} Nevertheless, most existing approaches either rely on comparatively large datasets, emphasize candidate generation without high-fidelity electronic-structure validation, or assess success primarily via aggregate distributions and a small set of representative examples. As a result, it remains unclear how reliably conditional control transfers to optical properties evaluated with \ac{tddft} in realistic low-data settings, and whether that reliability is consistent across chemical subspaces.

Here, we study conditional molecular generation for optical properties in a realistic low-data regime. We develop a token-conditioned \ac{gpt2} pipeline in which absorption energy and oscillator strength are discretized into property tokens and prepended to \ac{smiles} strings, while the \ac{homo-lumo} gap is included as an auxiliary electronic descriptor. The model is pretrained on a broad chemical corpus, trained on a large computational OLED dataset, and fine-tuned with multi-task optimization on a smaller curated set of molecules with consistent optical labels.
We then evaluate the generated candidates at the time-dependent density functional theory (TD-DFT) level, so that conditional success is assessed in terms of electronic-structure observables rather than token compliance or training-set proximity.

We benchmark the model at three complementary levels. 
First, we examine global distributional fidelity and show that the generated library reproduces the dominant optical-property support of the training data, spanning emission from the near-\ac{ir} to the \ac{uv} range, while shifting toward more compact structures, i.e. lower molecular weight and fewer heavy atoms.
Second, we quantify token-level controllability and find that conditioning is clearly directional but not fully orthogonal: steering one optical property influences the other, producing calibration errors across the target grid. 
Third, we carry out a chemotype-resolved analysis based on local electronic environments and show that controllability is strongly motif-dependent. Moderately conjugated aromatic-carbon environments are associated with improved joint target satisfaction, whereas electron-withdrawing motifs, particularly aryl nitriles, exhibit systematic red-shifting and reduced reliability. 
Together, these results show that token-level optical preconditioning is effective at a coarse level, but that its reliability ultimately depends on the local electronic environments through which the requested property shifts are realized.

\section{Results and discussion}
\begin{figure}[t]
    \centering
    \includegraphics[width=.95\textwidth]{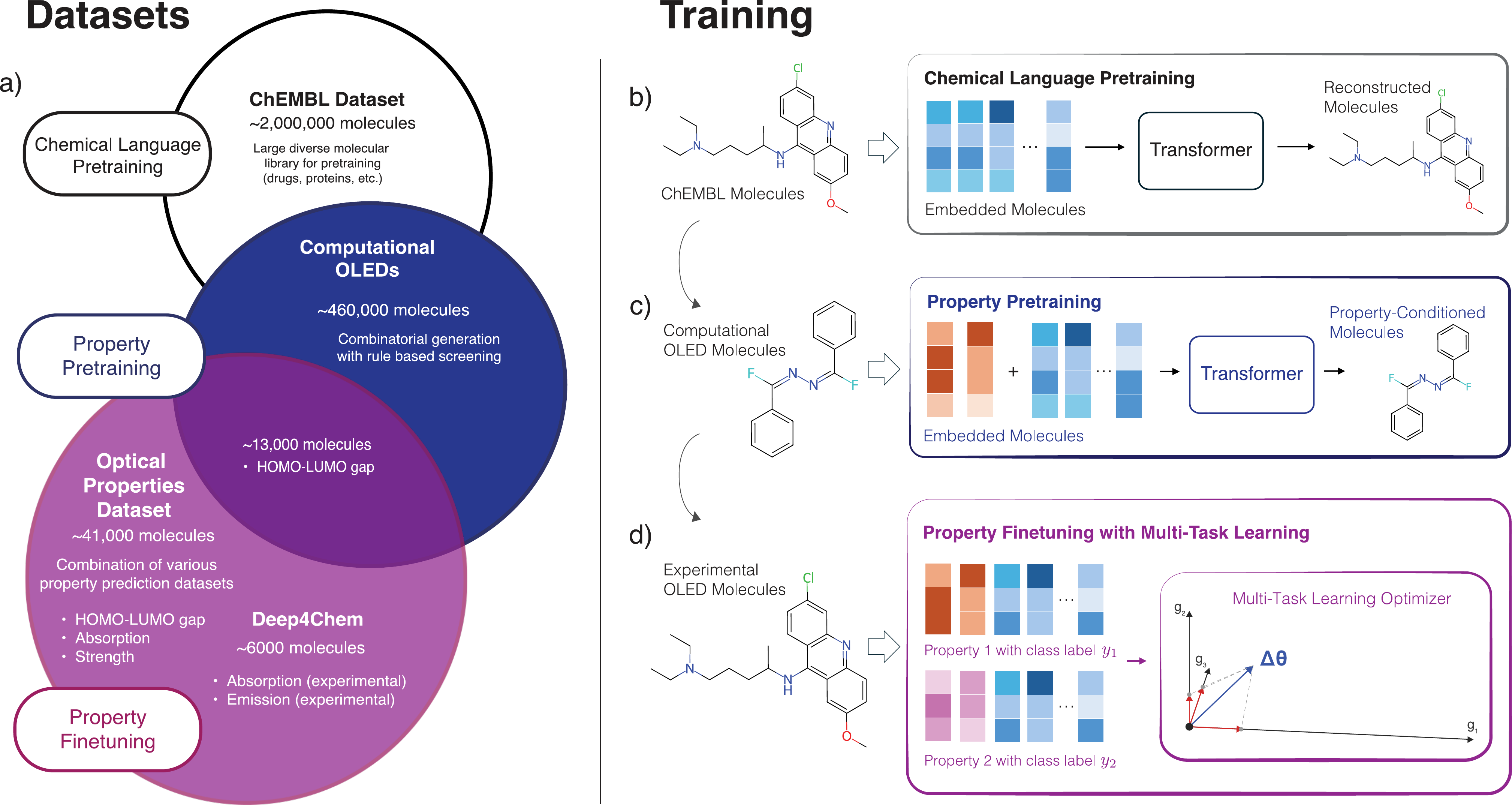}
    \caption{\textbf{Overview of the data sources and training pipeline.} \textbf{a)}~Details the properties of datasets used in each training phase. \textbf{b)}~Chemical language pretraining on the ChEMBL library, where the transformer learns valid, unconditional molecular generation. \textbf{c)}~Property pretraining on the computational OLED dataset, where absorption, oscillator strength, and HOMO--LUMO gap labels are introduced so the model learns property-token/structure associations. \textbf{d)}~Property finetuning on experimental data, where multi-task learning improves absorption- and intensity-conditioned generation. A multi-task learning optimizer guides model updates to mitigate data imbalance by optimizing the update vector $\Delta\theta$ to consider unnormalized gradients from each property class, e.g., $g_1$, $g_2$, $g_3$ equally. }
    \label{fig:oled_schematic}
\end{figure}

To investigate the efficacy of our property-conditioned generative model, we follow the pipeline illustrated in \cref{fig:oled_schematic}:
 Pretrain a language model on a large molecular corpus, introduce property conditioning on computed OLED data, and finetune on experimental molecules. We then generate candidate OLED molecules conditioned on desired property ranges and evaluate them with ab initio optical property calculations (\cref{sec:optical}).

Our analysis proceeds from model construction to electronic-structure validation of the generated molecules.%
We first describe the staged training strategy used to obtain a property-conditioned molecular generator (\cref{fig:oled_schematic}), which combines chemical-language pretraining, OLED-property pretraining, and multi-task fine-tuning on a curated low-data optical dataset.%
We then evaluate whether the resulting absorption-energy and oscillator-strength tokens provide genuine control over TD-DFT optical observables, rather than merely producing valid SMILES or molecules close to the training distribution.%
Finally, we examine whether this control is uniform across chemical space or instead depends on specific local electronic environments.

\subsection{Model architecture and training}
\begin{figure}[t]
    \centering
    \includegraphics[width=.80\textwidth]{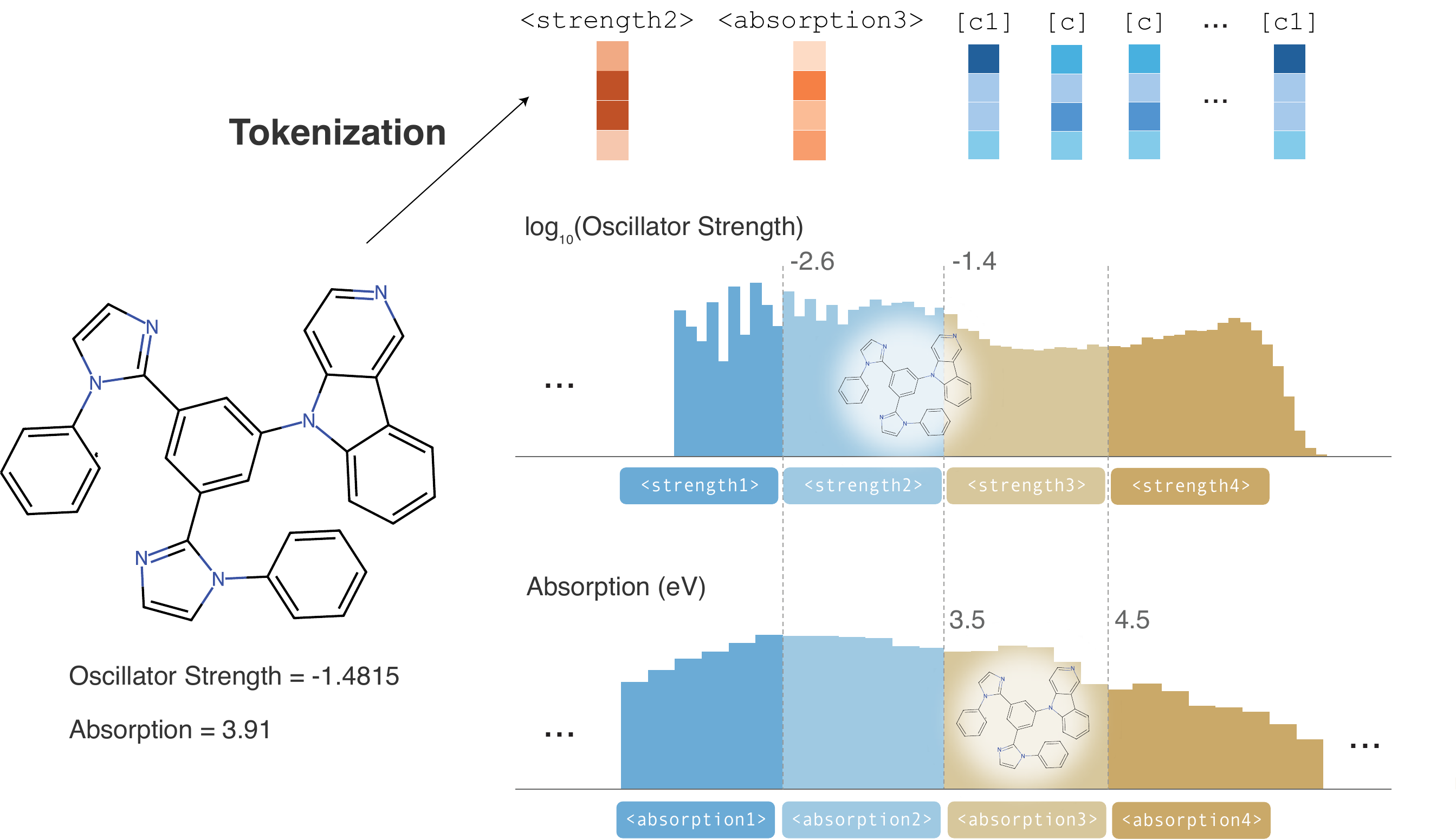}
    \caption{
    \textbf{Property-token embedding and molecular representation}. Continuous property values (e.g., oscillator strength and absorption energy) are converted into tokens via quantile-based discretization and prepended to the SMILES sequence as property tokens. Positional embeddings are zeroed at property-token positions so that the ordering of property tokens (when multiple properties are specified) does not affect the model's predictions.}
    \label{fig:ppty_binning}
\end{figure}

We use GPT-2,\cite{radford2019language} a decoder-only transformer, to autoregressively generate SMILES tokens, one by one, following the same broad autoregressive modelling paradigm that has also recently been extended to direct 3D molecular generation.\cite{chengScalableAutoregressive3DMolecule2025} Molecules are tokenized with the SMILES tokenizer of Ref. \citenum{schwaller_molecular_2019}, augmented with property tokens and sequence-control tokens. To enable property-conditioned generation, each molecule's computed absorption energy, oscillator strength, and HOMO--LUMO gap are discretized into bins and prepended to the SMILES sequence as special tokens (\cref{fig:ppty_binning}). Positional embeddings are zeroed at these positions so they function as global conditioning signals rather than sequential elements.\label{pretrain}\label{ARM}\label{ppt}\label{conditional}

Training proceeds in three stages (\cref{fig:oled_schematic}): (i) language pretraining on ChEMBL\cite{gaultonChEMBLLargescaleBioactivity2012} to learn SMILES grammar via next-token prediction;\cite{daiSemisupervisedSequenceLearning2015,radford_improving_nodate} (ii) property pretraining on the 460{,}205-molecule computational OLED dataset,\cite{gomez-bombarelliDesignEfficientMolecular2016} introducing the discretized property tokens; and (iii) multi-task finetuning on the curated dataset (${\sim}41{,}500$ molecules), balancing five objectives with Nash Multi-Task Learning to prevent catastrophic forgetting and gradient domination.\cite{navonMultiTaskLearningBargaining2022} Full architecture, tokenization, training hyperparameters, the classifier-free-guidance-style conditional sampling,\cite{sanchez_stay_2023} and multi-task details are provided in Supplementary Note~\ref{si:model_details}.

\subsection{Conditioned generation through property-tokens}

We generated candidates by sampling conditioned SMILES across the full grid of vertical absorption energy ($\Delta E_{\mathrm{0\rightarrow S_1}}$) and oscillator strength ($f_{\mathrm{S0 \rightarrow S1}}$) tokens, followed by canonicalization and removal of invalid and duplicate strings before quantum-chemical evaluation. Before analyzing the TD-DFT results, we first characterized the generated library at the string and scaffold levels. Across the absorption/oscillator-strength grid, validity and uniqueness at the token level varied substantially, indicating that conditional calibration already affects generation quality before electronic-structure evaluation (\cref{fig:generation_metrics_token_heatmaps}). At the same time, scaffold diversity remained consistently high, and pairwise overlap between conditioning-specific valid unique sets was generally low (\cref{fig:generation_metrics_token_heatmaps,fig:overlap_token_groups}), showing that different prompts do not merely reproduce a common pool of molecules, but instead sample largely distinct regions of chemical space. The valid unique library was also dominated by neutral molecules (1803 molecules, 90.9\%), and every conditioning pair contained enough neutral candidates to construct a final subset of 20 molecules per condition for further TD-DFT screening (\cref{fig:formal_charge_token_pair_counts}).

\begin{figure}[!htbp]
    \centering    \includegraphics[width=0.78\linewidth]{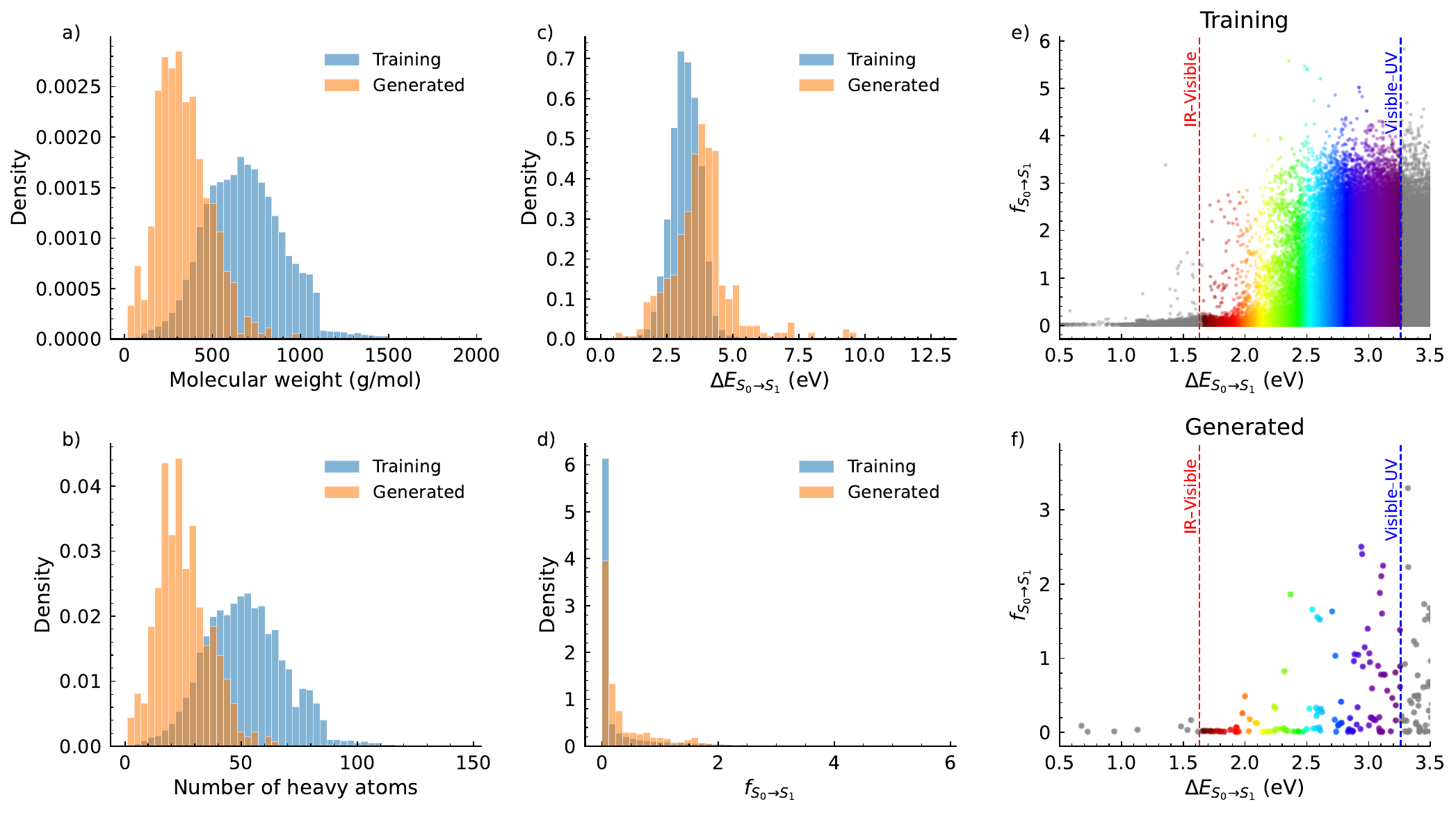}
  \caption{
\textbf{Global distributional fidelity and joint optical-property space of generated molecules.}
Density histograms compare the training and generated-molecules sets for
a) molecular weight, 
b) number of heavy atoms,
c) vertical $S_0 \rightarrow S_1$ absorption energy, and
d) oscillator strength.
The generated library reproduces the dominant absorption energy and oscillator strength support of the training data, while shifting toward more compact molecules with lower molecular weight and fewer heavy atoms.
e,f) Joint distributions of $\Delta E_{S_0\!\to\!S_1}$ and $f_{S_0\!\to\!S_1}$ for the training and generated sets, respectively.
Points are coloured by the wavelength corresponding to the excitation energy; grey points fall outside the plotted visible-colour wavelength range.
Dashed vertical lines indicate the approximate IR--visible and visible--UV boundaries at 1.63 and 3.26 eV.
The generated molecules mainly populate the visible region, with a sparse low-energy tail and generally reduced oscillator strengths near the infrared boundary.
Generated-molecule optical properties correspond to the 500 neutral candidates evaluated with the TD-DFT validation workflow described in Methods.
}
\label{fig:training_generated_distributions_and_scatter}

\end{figure}

At the level of global property distributions, the final generated library reproduces the dominant absorption-energy and oscillator-strength support of the training set while shifting toward lower molecular weight and fewer heavy atoms (\cref{fig:training_generated_distributions_and_scatter}a--d). This behaviour indicates that the model remains largely on the optical-property manifold represented in the training data, yet prefers to explore more compact molecular scaffolds. The generated set, therefore, preserves the main property support while shifting toward a structurally distinct region of chemical space, rather than merely reproducing the training distribution.

A complementary view is provided by the joint TD-DFT property distribution in \cref{fig:training_generated_distributions_and_scatter}, panels e and f. The generated molecules densely populate the visible regime and extend into a smaller, low-energy subset near the infrared regime and, in a few cases, within it. This narrow-gap tail is not the dominant outcome of generation. Even so, it is chemically informative because it shows that the model can access electronically distinct solutions beyond the main visible-region support. Notably, oscillator strengths in this low-gap regime are generally reduced, indicating that red-shifted candidates are not obtained with uniformly strong transition intensities. Representative infrared candidates shown in Fig.~\ref{fig:ir_molecules_SA} in the Supporting Information further suggest that the more synthetically accessible members of this subset are typically based on extended fused or non-benzenoid $\pi$-frameworks, whereas the least accessible examples rely on more elaborate structural motifs. We therefore interpret the infrared tail as a plausible but comparatively sparse frontier of the learned distribution.

\begin{figure}[t]
    \centering
    \includegraphics[width=0.8\linewidth]{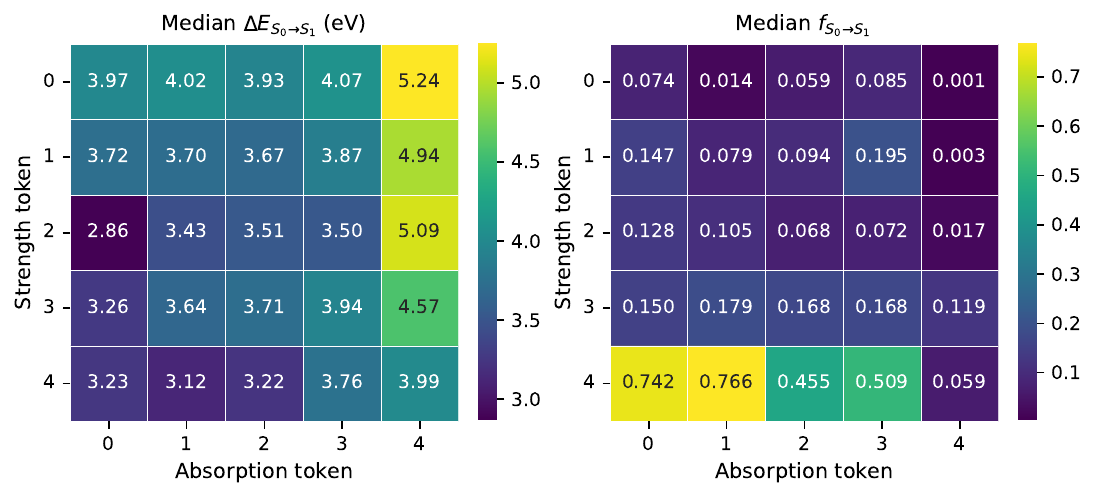}
    \caption{
Token-conditioned property control evaluated at the TD-DFT level. 
Heatmaps report the median vertical absorption energy 
$\Delta E_{S_0\!\to\!S_1}$ (left, in eV) and median oscillator strength 
$f_{S_0\!\to\!S_1}$ (right) as a function of the conditioning bins 
for absorption (x-axis) and strength (y-axis). 
Each cell corresponds to the median value computed over all generated 
molecules assigned to the corresponding $(\text{absorption bin}, 
\text{strength bin})$ pair. 
Overall directional trends are clear, but local deviations (including a strength-bin inversion and cross-axis coupling) indicate partial non-orthogonality in control.
}
    \label{fig:median_heatmaps_td-dft}
\end{figure}

To quantify controllability directly, we next examined the TD-DFT outcomes for each conditioning bin using the median statistics in \cref{tab:median_properties} and the full heatmaps in \cref{fig:median_heatmaps_td-dft}. The median absorption energy increases from 3.469 to 4.601 eV across vertical absorption bins, indicating clear directional steering along the excitation-energy axis. The median oscillator strength also increases overall from 0.043 to 0.465 across strength bins, although this trend is not strictly monotonic because strength bin 2 falls below bin 1. Thus, token conditioning is effective at a coarse level, but not perfectly calibrated.

The heatmaps additionally show that conditioning on vertical absorption affects oscillator strength, and vice versa. Although non-monotonically, these two properties are anti-correlated. This coupling is consistent with the energy gap law: as absorption shifts to lower energies, the reduced HOMO--LUMO gap tends to suppress radiative decay rates and thus oscillator strength. Note that this is a distinct claim from the joint distribution shown in \cref{fig:training_generated_distributions_and_scatter}e,f, which reflects the spread of individual molecules across property space. The question here is whether the \emph{bin-level medians} in \cref{fig:median_heatmaps_td-dft} reproduce the energy gap law trend across the conditioning grid. To quantify this, we computed the Spearman rank correlation between the marginal median absorption energy and marginal median oscillator strength across conditioning bins (Table~S1), using the curated $\omega$B97X-D3 fine-tuning set as the reference for the training comparison (the like-for-like level to the TD-DFT validation). The training-set correlation is weak and non-significant ($\rho = 0.235$, $p = 0.36$, 17 bins), which is consistent with the relatively flat upper envelope of oscillator strength across the visible range: the energy gap law constraint becomes most pronounced only below approximately 2.0~eV, a regime sparsely represented in the training data (\cref{fig:median_heatmaps_td-dft}a). By contrast, the generated-set medians show a stronger and statistically significant correlation ($\rho = 0.650$, $p = 0.022$, 12 bins), indicating that the conditioning grid amplifies the anti-correlation relative to what is present in the training distribution. A Fisher $z$-test comparing the two Spearman coefficients does not reach significance ($z = -1.255$, $p = 0.21$), so this difference should be interpreted cautiously. The directional pattern is nevertheless consistent with the generated molecules being preferentially drawn from the lower-energy, lower-oscillator-strength frontier of the learned distribution, as is also visible in the sparse infrared tail of \cref{fig:training_generated_distributions_and_scatter}f.

Taken together, these results establish two complementary points. First, the model generates a diverse and predominantly neutral candidate library that remains close to the dominant optical-property support of the training data while exploring structurally more compact molecules. Second, this global fidelity coexists with an uneven controllability landscape, in which some target regions are more reliably reached than others. This heterogeneity motivates the chemotype-resolved analysis in \cref{sec:chemotype}, where we examine which local electronic environments are associated with robust conditional control and which systematically bias the generated molecules away from the requested targets.

\subsection{Chemotype-Dependent Reliability and the Role of Local Electronic Environments}
\label{sec:chemotype}

To determine whether conditional generation reliability depends on interpretable, local chemical structure, we required a representation that can resolve local electronic environments rather than coarse fragments or global descriptors. Fragment-based schemes, such as BRICS, preserve synthetic building blocks but discard electronic context, whereas global descriptor sets, such as Mordred, obscure structural locality. Because local orbital interactions and substituent effects govern conditional control of optical properties, we employed the Overlapping Fragment Molecular Representation (OFraMP),\cite{stroetOFraMPFragmentbasedTool2023} which encodes atom-centred subgraphs at fixed graph radii and retains chemically meaningful local electronic environments.

Stable OFraMP environments (support $\geq 30$ molecules) were grouped into motif classes defined by radius, central atom identity, and coarse chemical context (e.g., aromatic carbon, nitrogen-centred, carbonyl-like). For each motif class, we computed the conditional joint success rate for absorption- and oscillator-strength-bin targeting and compared it to the global baseline ($\approx 10\%$). Many highlighted classes remain in the modest-support regime ($n\approx 30$--$86$), so effect sizes should be interpreted as chemotype-level signals rather than definitive rankings.

\begin{figure}[H]
\centering
\setlength{\tabcolsep}{3pt}
\renewcommand{\arraystretch}{1.0}
\begin{tabular}{@{}cc@{}}
\begin{minipage}[t]{0.48\linewidth}
\centering
Enriched motif: \texttt{r2::C::Aromatic}\\
\footnotesize Radius-$2$ neighbourhood; core atom = C; aromatic local environment\\
\footnotesize $n=30$ molecules, success rate $=0.27$, lift $=2.66\times$ baseline\\[0.4em]
\begin{tabular}{@{}cc@{}}
\includegraphics[width=0.45\linewidth]{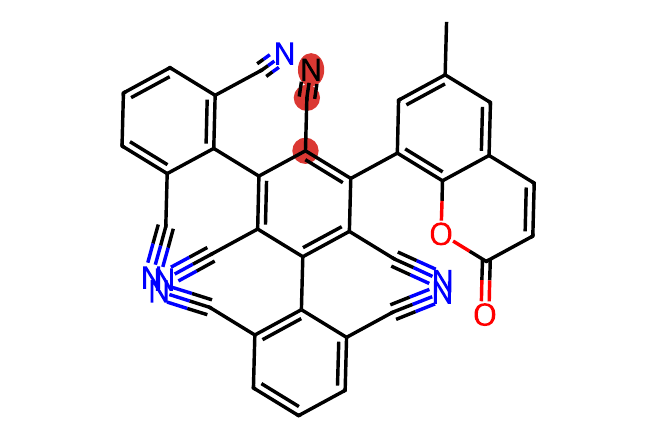}
 & 
\includegraphics[width=0.45\linewidth]{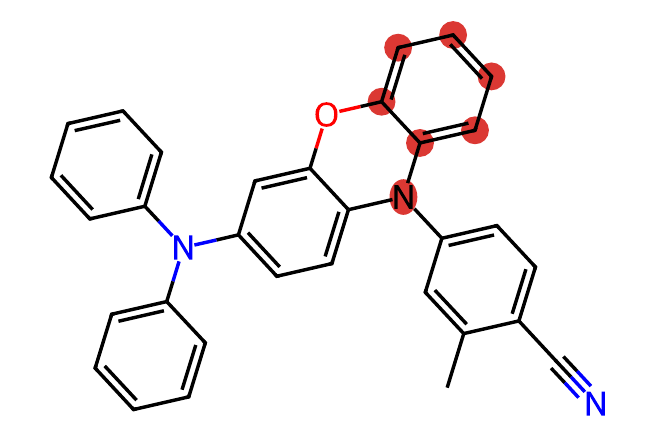}
\\[0.4em]
\includegraphics[width=0.45\linewidth]{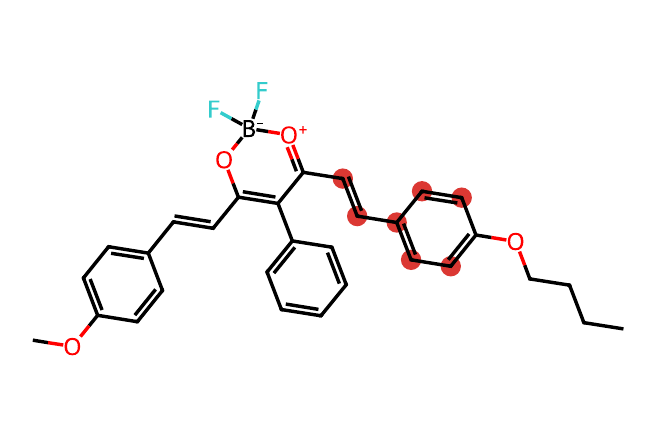}
 & 
\includegraphics[width=0.45\linewidth]{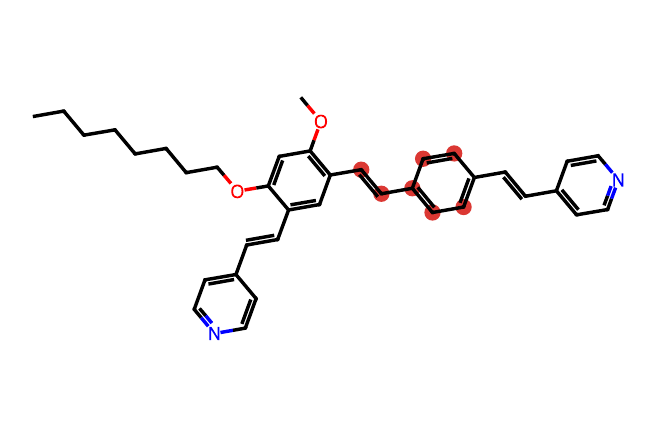}
\\[0.4em]
\end{tabular}
\end{minipage}
&
\begin{minipage}[t]{0.48\linewidth}
\centering
Enriched motif: \texttt{r1::O::Aliphatic}\\
\footnotesize Radius-$1$ neighbourhood; core atom = O; aliphatic local environment\\
\footnotesize $n=65$ molecules, success rate $=0.20$, lift $=1.99\times$ baseline\\[0.4em]
\begin{tabular}{@{}cc@{}}
\includegraphics[width=0.45\linewidth]{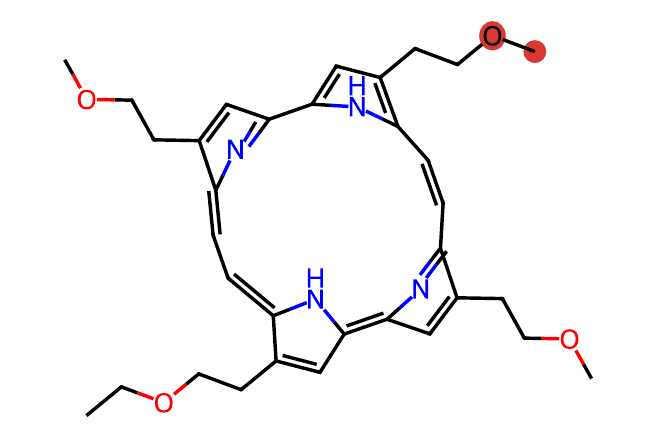}
 & 
\includegraphics[width=0.45\linewidth]{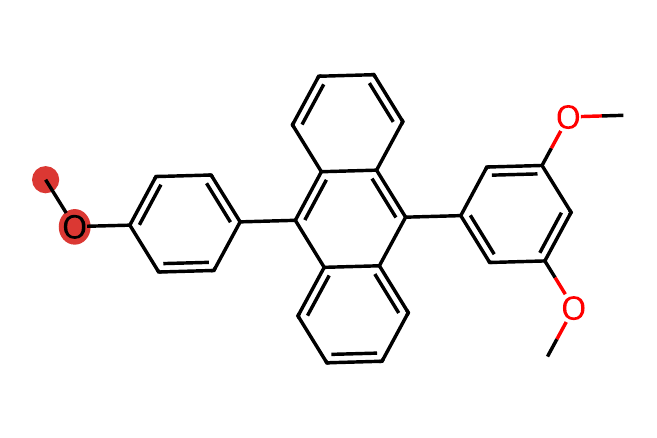}
\\[0.4em]
\includegraphics[width=0.45\linewidth]{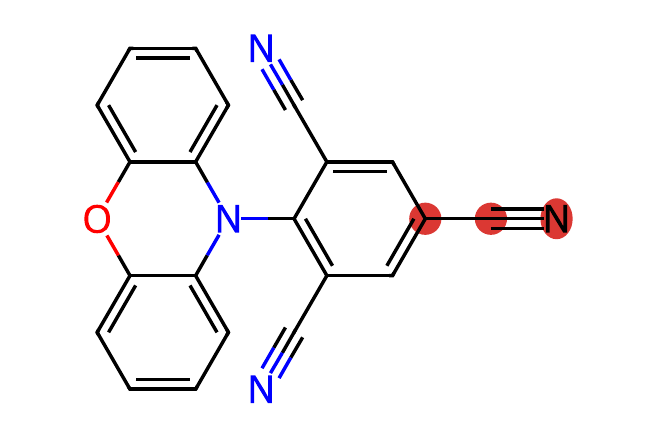}
 & 
\includegraphics[width=0.45\linewidth]{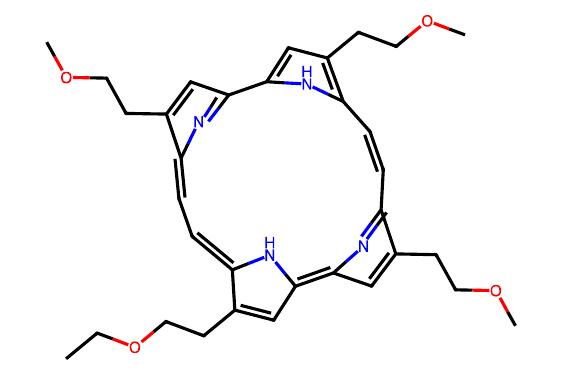}
\\[0.4em]
\end{tabular}
\end{minipage}\\[1.2em]
\begin{minipage}[t]{0.48\linewidth}
\centering
Suppressed motif: \texttt{r2::cC::N\_env}\\
\footnotesize Radius-$2$ neighbourhood; core atom pair = aromatic c--C; nitrogen in the surrounding environment\\
\footnotesize $n=32$ molecules, success rate $=0.00$, lift $=0.00\times$ baseline\\[0.4em]
\begin{tabular}{@{}cc@{}}
\includegraphics[width=0.45\linewidth]{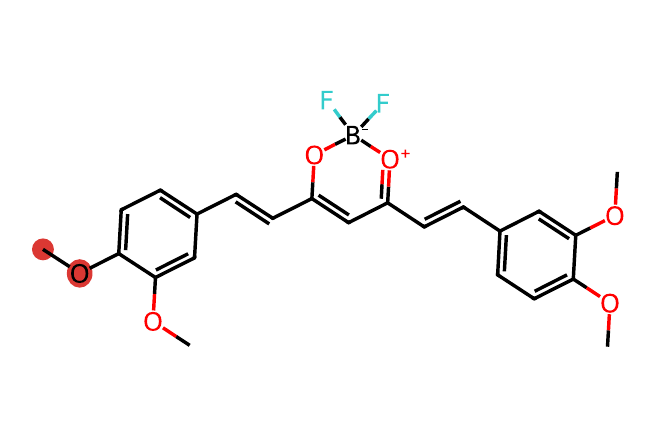}
 & 
\includegraphics[width=0.45\linewidth]{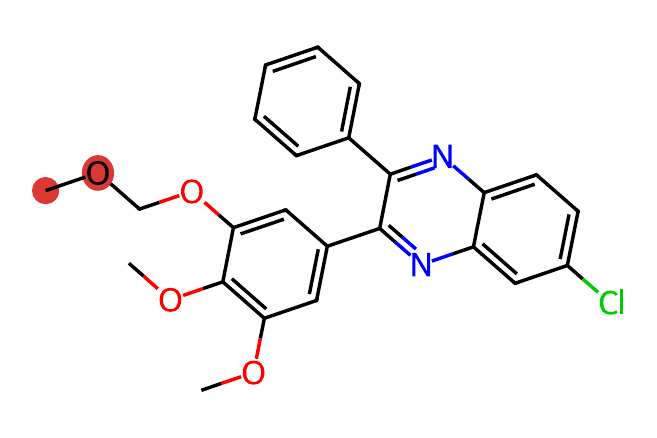}
\\[0.4em]
\includegraphics[width=0.45\linewidth]{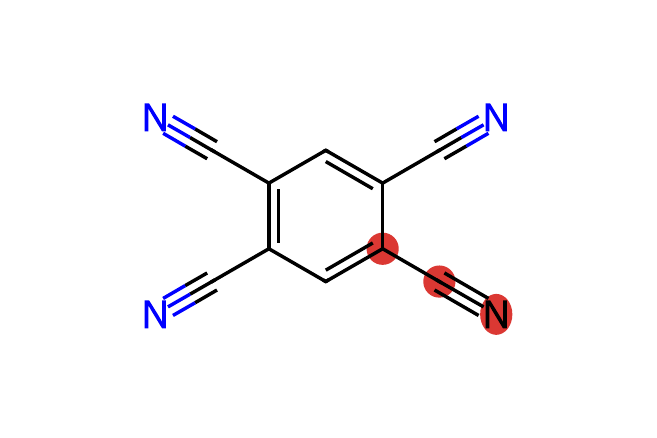}
 & 
\includegraphics[width=0.45\linewidth]{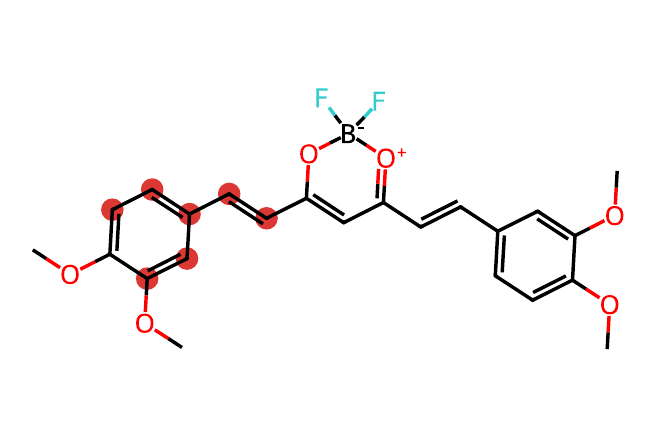}
\\[0.4em]
\end{tabular}
\end{minipage}
&
\begin{minipage}[t]{0.48\linewidth}
\centering
Suppressed motif: \texttt{r3::Nc::Aromatic}\\
\footnotesize Radius-$3$ neighbourhood; core atom pair = N--aromatic c; aromatic local environment\\
\footnotesize $n=86$ molecules, success rate $=0.03$, lift $=0.35\times$ baseline\\[0.4em]
\begin{tabular}{@{}cc@{}}
\includegraphics[width=0.45\linewidth]{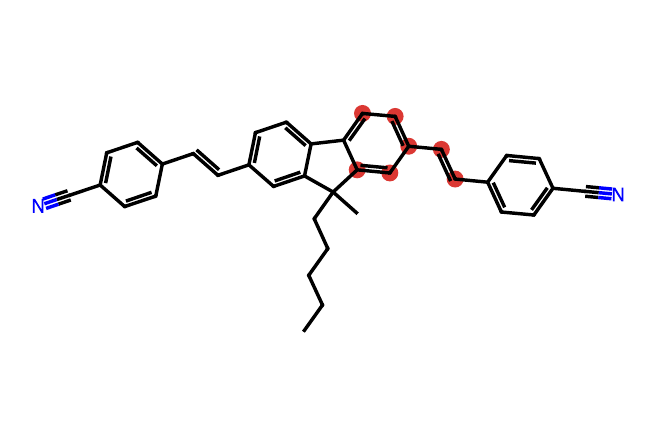}
 & 
\includegraphics[width=0.45\linewidth]{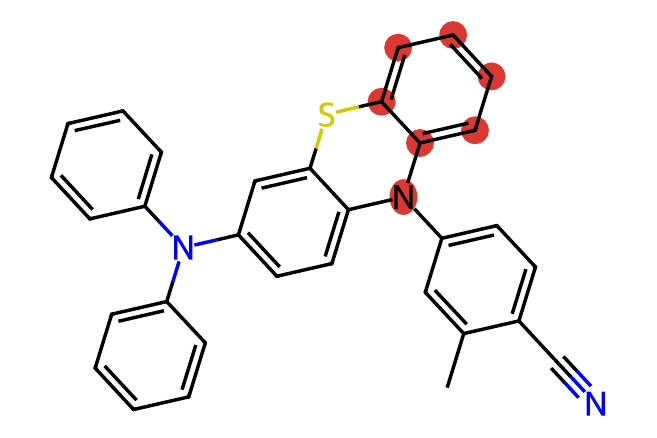}
\\[0.4em]
\includegraphics[width=0.45\linewidth]{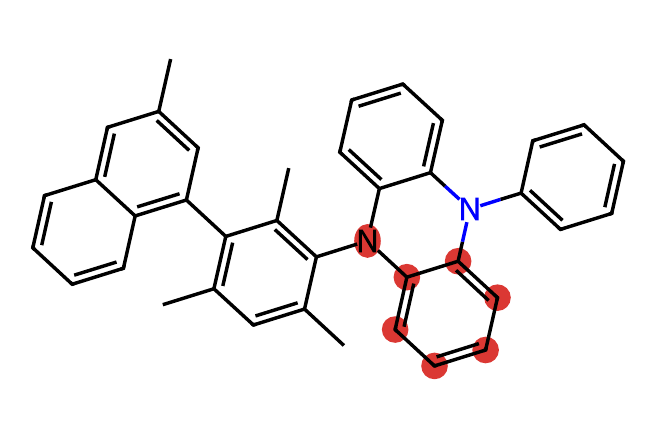}
 & 
\includegraphics[width=0.45\linewidth]{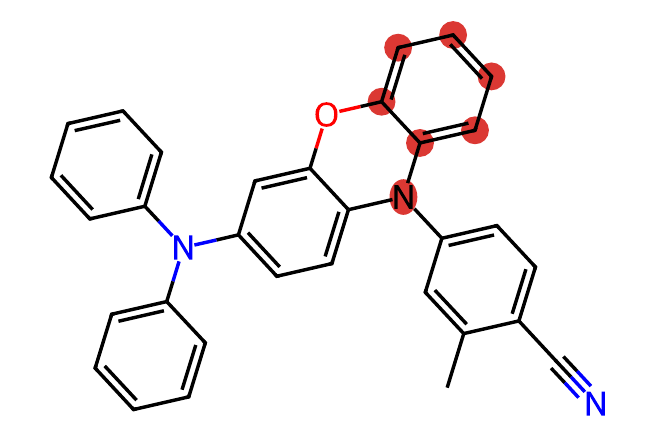}
\\[0.4em]
\end{tabular}
\end{minipage}\\[1.2em]
\end{tabular}
\caption{\textbf{Structure-level exemplars of enriched and suppressed chemotypes.} Each panel is labelled with the exact OFraMP motif code and a plain-language translation. Here, $r$ denotes the radius of the local neighbourhood used to define the motif, the \emph{core} is the central atom or atom pair in that motif, $n$ is the number of generated molecules containing the motif, \emph{success rate} is the fraction of those molecules that satisfy both the target absorption and oscillator-strength bins, and \emph{lift} is the success-rate enrichment relative to the overall baseline. The top row shows enriched motifs with above-baseline controllability, whereas the bottom row shows suppressed motifs. For motifs labelled \texttt{N\_env}, nitrogen is present somewhere within the radius-$r$ neighbourhood, even if the highlighted core atom(s) are carbon. Throughout, absorption bins run from 0 (red-shifted / lower energy) to 4 (blue-shifted / higher energy), and strength bins run from 0 (weak) to 4 (strong), following the convention defined in \cref{fig:chemotype_pairs}.}
\label{fig:chemotype_examples}
\end{figure}

\begin{figure}[H]
\centering
\setlength{\tabcolsep}{1pt}
\begin{minipage}[t]{0.56\linewidth}
\centering
\textbf{Matched success/failure pairs}\\
\footnotesize Coordinates are reported as (absorption bin, strength bin); absorption runs from 0 = red-shifted / lower energy to 4 = blue-shifted / higher energy, and strength runs from 0 = weak to 4 = strong.\\[0.4em]
\begin{tabular}{@{}cc@{}}
\begin{minipage}[t]{0.475\linewidth}
\centering
\includegraphics[width=\linewidth]{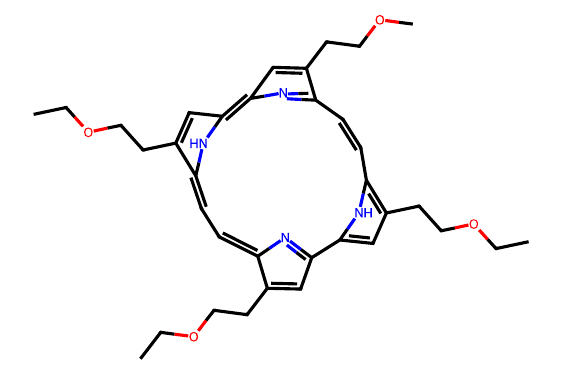}\\[-0.2em]
\scriptsize \textbf{Success} $\;$ target (abs 0, str 3), achieved (abs 0, str 3)\\
\scriptsize $\Delta$abs +0, $\Delta f$ +0
\end{minipage}
&
\begin{minipage}[t]{0.475\linewidth}
\centering
\includegraphics[width=\linewidth]{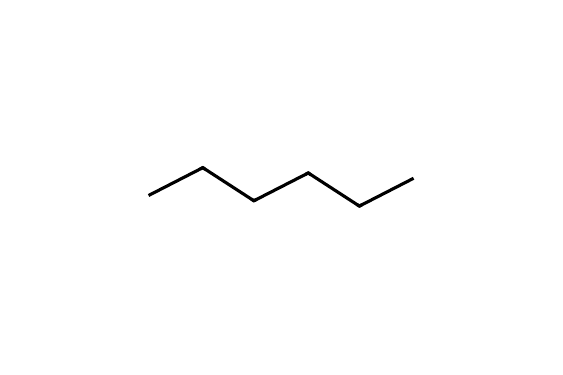}\\[-0.2em]
\scriptsize \textbf{Failure} $\;$ target (abs 0, str 3), achieved (abs 0, str 4)\\
\scriptsize $\Delta$abs +0, $\Delta f$ +1
\end{minipage}
\\[0.8em]
\begin{minipage}[t]{0.475\linewidth}
\centering
\includegraphics[width=\linewidth]{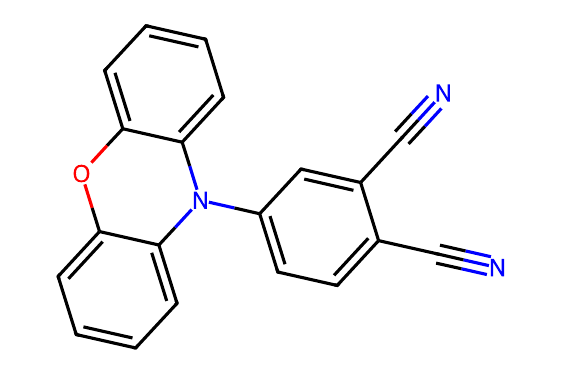}\\[-0.2em]
\scriptsize \textbf{Success} $\;$ target (abs 4, str 0), achieved (abs 4, str 0)\\
\scriptsize $\Delta$abs +0, $\Delta f$ +0
\end{minipage}
&
\begin{minipage}[t]{0.475\linewidth}
\centering
\includegraphics[width=\linewidth]{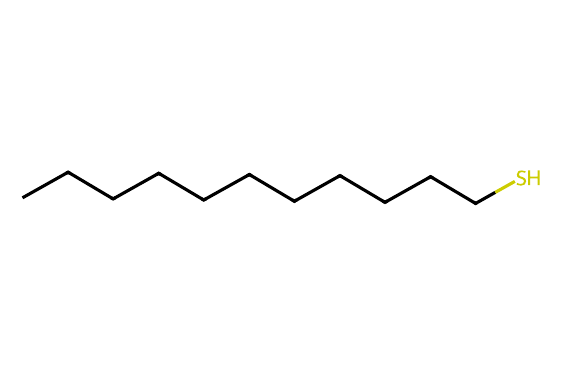}\\[-0.2em]
\scriptsize \textbf{Failure} $\;$ target (abs 4, str 0), achieved (abs 4, str 1)\\
\scriptsize $\Delta$abs +0, $\Delta f$ +1
\end{minipage}
\\[0.8em]
\begin{minipage}[t]{0.475\linewidth}
\centering
\includegraphics[width=\linewidth]{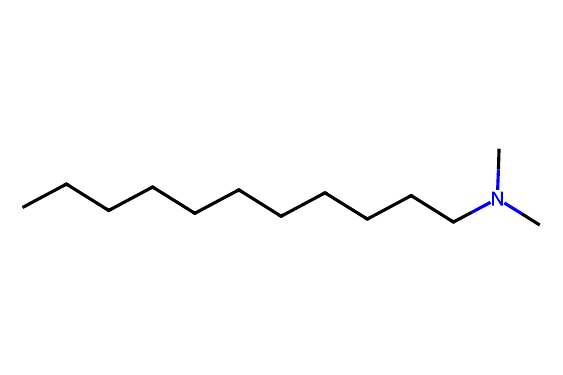}\\[-0.2em]
\scriptsize \textbf{Success} $\;$ target (abs 4, str 1), achieved (abs 4, str 1)\\
\scriptsize $\Delta$abs +0, $\Delta f$ +0
\end{minipage}
&
\begin{minipage}[t]{0.475\linewidth}
\centering
\includegraphics[width=\linewidth]{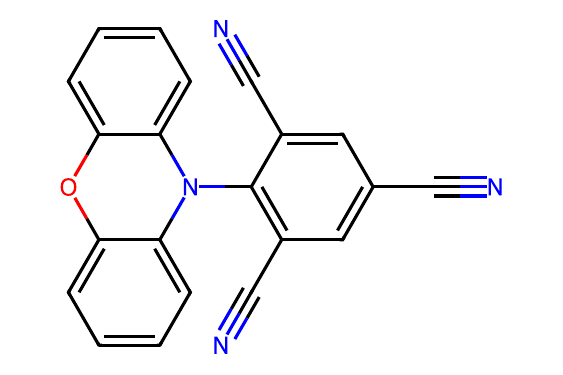}\\[-0.2em]
\scriptsize \textbf{Failure} $\;$ target (abs 4, str 1), achieved (abs 4, str 3)\\
\scriptsize $\Delta$abs +0, $\Delta f$ +2
\end{minipage}
\\[0.8em]
\end{tabular}
\end{minipage}
\hfill
\begin{minipage}[t]{0.40\linewidth}
\begin{minipage}[t]{\linewidth}
\centering
\textbf{Analog series}\\
\footnotesize Small structural edits can move molecules from lower- to higher-index bins and reveal where calibration breaks down.\\[0.3em]
\begin{tabular}{@{}c@{}}
\begin{minipage}[t]{0.92\linewidth}
\centering
\includegraphics[width=0.70\linewidth]{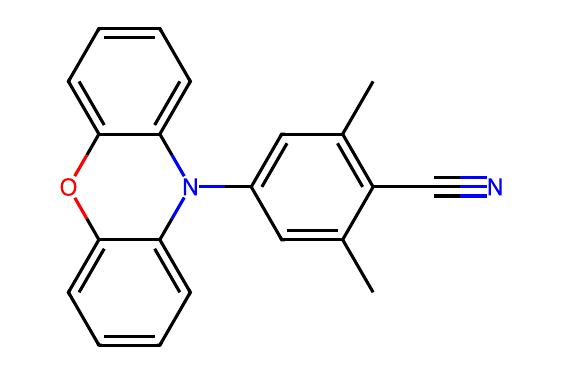}\\[-0.2em]
\scriptsize target (abs 0, str 3), achieved (abs 1, str 3)\\
\scriptsize $\Delta$abs +1, $\Delta f$ +0
\end{minipage}\\[0.7em]
\begin{minipage}[t]{0.92\linewidth}
\centering
\includegraphics[width=0.70\linewidth]{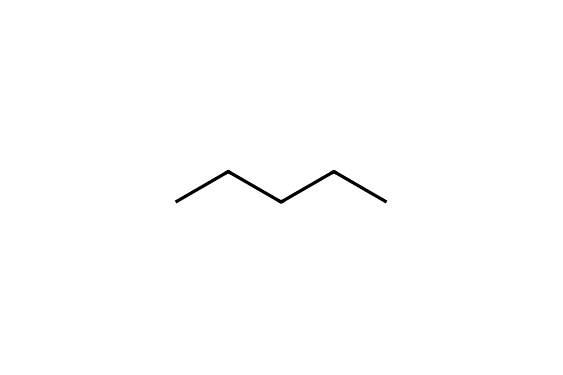}\\[-0.2em]
\scriptsize target (abs 2, str 2), achieved (abs 3, str 3)\\
\scriptsize $\Delta$abs +1, $\Delta f$ +1
\end{minipage}\\[0.7em]
\begin{minipage}[t]{0.92\linewidth}
\centering
\includegraphics[width=0.70\linewidth]{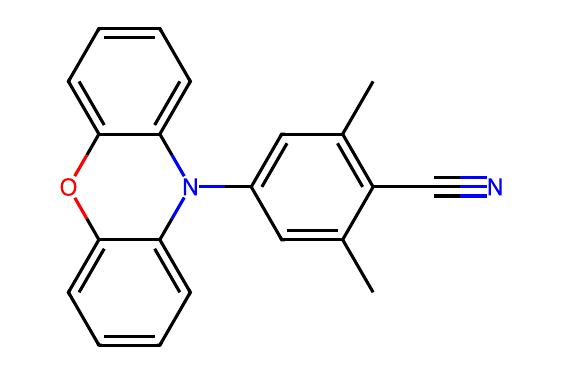}\\[-0.2em]
\scriptsize target (abs 0, str 0), achieved (abs 4, str 3)\\
\scriptsize $\Delta$abs +4, $\Delta f$ +3
\end{minipage}
\end{tabular}
\end{minipage}
\end{minipage}
\caption{\textbf{Matched comparisons and analog drift reveal failure mechanisms.} Left: matched pairs drawn from the same requested conditioning bins, where \emph{success} means that the TD-DFT-evaluated molecule lands in both requested bins and \emph{failure} means that at least one property misses its target. Right: a short analog series showing that modest structural edits can move a molecule across progressively higher absorption and oscillator-strength bins. Throughout, the first coordinate is the absorption bin and the second is the oscillator-strength bin; lower absorption-bin indices correspond to red-shifted / lower-energy excitations, whereas higher indices correspond to blue-shifted / higher-energy excitations.}
\label{fig:chemotype_pairs}
\end{figure}

To visually connect these statistics to concrete chemistry, we include two structure-centric figures. Figure~\ref{fig:chemotype_examples} summarizes representative molecules for enriched and suppressed motif classes with highlighted local OFraMP environments and explicitly defines the motif notation ($r$, core identity, $n$, success rate, and lift). Figure~\ref{fig:chemotype_pairs} then presents matched success/failure comparisons and short analog series, using the coordinate convention (absorption bin, strength bin), where absorption runs from bin 0 (red-shifted / lower energy) to bin 4 (blue-shifted / higher energy) and strength runs from bin 0 (weak) to bin 4 (strong).

\subsubsection*{Enriched Motif Classes}

Table~\ref{tab:motif_top} lists the most enriched motif classes ranked by lift relative to baseline.

\begin{table}[ht]
\centering
\caption{\textbf{Top motif classes ranked by reliability lift (support $\geq 30$).}}
\label{tab:motif_top}
\begin{tabular}{lccc}
\hline
Motif Class & $n$ & Success Rate & Lift \\
\hline
r2::C::Aromatic & 30 & 0.267 & 2.65 \\
r1::O::Aliphatic & 65 & 0.200 & 1.99 \\
r1::C::Carbonyl\_like & 30 & 0.167 & 1.66 \\
r2::CC::O\_env & 30 & 0.167 & 1.66 \\
r3::Cc::Aromatic & 50 & 0.160 & 1.59 \\
\hline
\end{tabular}
\end{table}

The most enriched class, \texttt{r2::C::Aromatic}, corresponds to radius-2 aromatic carbon environments embedded within moderately conjugated $\pi$ systems. Molecules containing this motif achieved a joint success rate of $26.7\%$, corresponding to a $2.65$-fold enrichment relative to baseline. Absorption bin deviations for this class were centred near zero (13/30 exact matches), with modest symmetric dispersion. These environments therefore appear to occupy a smoothly tunable electronic regime in which token-level conditioning can modulate the HOMO--LUMO gap and oscillator strength within bin resolution.

Several oxygen-containing and carbonyl-like motifs also exhibited moderate enrichment (lift $\approx 1.6$--$2.0$), indicating that not all polar substituents degrade controllability. Rather, moderate electronic perturbations remain within the calibration range of the discrete conditioning tokens.

\subsubsection*{Suppressed Motif Classes}

In contrast, several nitrogen-centred and extended aromatic motifs exhibited suppressed reliability (Table~\ref{tab:motif_bottom}).

\begin{table}[ht]
\centering
\caption{\textbf{Bottom motif classes ranked by lift (support $\geq 30$).}}
\label{tab:motif_bottom}
\begin{tabular}{lccc}
\hline
Motif Class & $n$ & Success Rate & Lift \\
\hline
r2::cC::N\_env & 32 & 0.000 & 0.00 \\
r3::Nc::Aromatic & 86 & 0.035 & 0.35 \\
r2::cc::N\_env & 130 & 0.038 & 0.38 \\
r3::ccc::Aromatic & 95 & 0.053 & 0.52 \\
r3::cc::Aromatic & 154 & 0.058 & 0.58 \\
\hline
\end{tabular}
\end{table}

The most striking case was the aryl nitrile environment \texttt{ENV[r=2]::cC\# [N:1]}, for which zero joint successes were observed among 32 molecules. Absorption bin deviations for this motif were strongly skewed toward positive shifts, with a mean bin shift of $+1.75$ relative to the requested target. By contrast, \texttt{r2::C::Aromatic} environments exhibited a mean shift of $+0.90$. The difference was statistically significant (Mann--Whitney $p = 0.031$; Cohen's $d = 0.51$), indicating a moderate increase in red-shifting bias associated with the nitrile motif; however, given the modest sample sizes, this comparison should be interpreted as exploratory rather than definitive. Oscillator strength deviations were relatively symmetric and centred near zero for both motifs, suggesting that absorption miscalibration is the dominant failure mechanism.

Other nitrogen-centred aromatic environments exhibited similar suppression (lift $\approx 0.35$--$0.38$), indicating that strong heteroatom-centred donor or acceptor motifs introduce electronic shifts exceeding the resolution of discrete bin conditioning.

\FloatBarrier

\FloatBarrier

\subsubsection*{Connection to the Energy Gap Law}\label{energy_gap_law}

The observed motif-dependent red-shifting bias can be interpreted in the context of the energy gap law. Strong electron-withdrawing motifs such as aryl nitriles stabilize the LUMO and reduce the HOMO--LUMO gap, thereby shifting absorption to longer wavelengths. In regimes where the optical gap decreases substantially, radiative and nonradiative decay rates become increasingly sensitive to small changes in electronic structure. The energy gap law predicts enhanced nonradiative decay as the gap narrows, imposing intrinsic constraints on achievable oscillator strength and efficiency at long wavelengths.

In this context, discrete bin conditioning operates at a finite resolution in property space, while strong acceptor motifs induce electronic shifts whose magnitude can exceed that resolution. For nitrile-containing systems, the systematic $+1.75$ bin absorption overshoot indicates that local electronic perturbations drive the molecule into a red-shifted regime beyond the intended target. Because joint success requires simultaneous satisfaction of the absorption and oscillator-strength bins, even a moderate systematic overshoot sharply reduces the probability of joint compliance.

By contrast, moderately conjugated aromatic carbon environments introduce smaller, smoother perturbations to the electronic structure. These systems remain within a quasi-linear response regime in which bin-level conditioning can effectively compensate for structural variation. Generic aromatic carbon backbones (e.g., \texttt{r0} and \texttt{r1} aromatic environments), which appear ubiquitously across the dataset, exhibit near-baseline performance, confirming that reliability is not determined by aromaticity per se but by specific heteroatom-centred motifs that amplify the electronic response.

Collectively, these results reveal a structured reliability landscape in conditional molecular generation. Rather than uniform stochastic failure, the model exhibits predictable degradation in chemical subspaces characterized by strong local acceptor motifs that induce large reductions in the gap. The interplay between discrete conditioning granularity and the magnitude of the motif-dependent electronic response provides a mechanistic explanation for the observed chemotype-dependent controllability.

\section{Conclusion}

We benchmark a token-conditioned autoregressive language model for generating OLED candidate molecules with targeted optical properties in a realistic low-data regime. Evaluated at the TD-DFT level, the generated molecules reproduce the dominant absorption energy and oscillator strength support of the training distribution while shifting toward lower molecular weight and fewer heavy atoms, indicating that conditional generation remains largely on-manifold while exploring structurally distinct regions of chemical space.

At the level of token control, the model shows clear directional steering across absorption-energy and oscillator-strength bins. Still, this control is not fully orthogonal and exhibits localized calibration irregularities. Therefore, discrete conditioning is effective at a coarse level, yet its reliability varies across property space and cannot be inferred from marginal distributions alone.

A central result of this study is that controllability is strongly chemotype-dependent. Local-environment analysis based on OFraMP reveals that moderately conjugated aromatic-carbon motifs are associated with improved joint target satisfaction. In contrast, electron-withdrawing motifs, particularly aryl nitriles, show systematic red-shifting and markedly reduced controllability. These findings support a mechanistic interpretation in which the success of discrete property conditioning depends on whether local electronic perturbations remain within the effective resolution of the conditioning scheme.

Taken together, our results recast conditional OLED molecular generation as a quantitative benchmarking problem rather than a purely generative one. They show that aggregate validity or distributional agreement is insufficient to assess model performance, and that chemically resolved analyses are required to identify where control is reliable, where it degrades, and how future models may be improved.

\section{Methods}

\subsection{Data and Representation}\label{sec:data}

Molecular structures were represented as SMILES strings, which allowed the generative task to be formulated as autoregressive sequence modelling with a GPT-2 architecture.\cite{radford2019language}

The data used for training follows the three-stage workflow summarised in \cref{fig:oled_schematic}: chemical-language pretraining, property pretraining, and final property fine-tuning.
Full details of the model architecture, tokenization, optimization procedure, and multi-task training strategy are provided in \cref{si:model_details}.

In the first stage, the model was pretrained on the ChEMBL molecular corpus.\cite{gaultonChEMBLLargescaleBioactivity2012}
This stage did not use optical-property labels; instead, it aimed to learn SMILES syntax, common molecular substructures, and a broad chemical prior via next-token prediction.
This chemical-language pretraining enables the base model to generate syntactically valid and chemically plausible molecular strings before introducing property conditioning.

In the second stage, property tokens were introduced using the large computational OLED dataset of Gómez-Bombarelli and co-workers.\cite{gomez-bombarelliDesignEfficientMolecular2016}
This virtual library contains 460{,}205 chromophores constructed from donor and acceptor fragments.
For these molecules, vertical absorption energies and oscillator strengths were computed with TD-DFT at the B3LYP/6-31G(d) level of theory.
For a subset of 12{,}867 molecules, HOMO and LUMO energies were computed directly; for the remaining molecules, frontier-orbital energies were estimated using a Chemprop directed message-passing neural network (D--MPNN) model.\cite{yangAnalyzingLearnedMolecular2019}%
During this property-pretraining stage, the computed absorption energy, oscillator strength, and HOMO--LUMO gap were discretized into property tokens and prepended to the SMILES sequence, enabling the model to learn associations between coarse optical-property regimes and molecular structure.

In the final stage, the model was fine-tuned on a smaller curated dataset designed to provide more consistent high-quality optical labels.%
This curated set combines two sources.%
The first source comprises 28{,}804 molecules from Greenman and co-workers,\cite{Greenman2022} for which absorption energies, oscillator strengths, and HOMO and LUMO energies were computed with TD-DFT at the $\omega$B97X-D3/def2-SVPD level.%
Although these molecules originate from experimental UV--Vis compilations --including CDEx,\cite{beardComparativeDatasetExperimental2019} ChemFluor,\cite{juMachineLearningEnables2021} Deep4Chem,\cite{joungExperimentalDatabaseOptical2020} DSSCDB,\cite{venkatramanDyesensitizedSolarCell2018} and DyeAgg\cite{venkatramanOpenAccessData2020} --the property tokens used in this work were derived from the consistently recomputed TD-DFT quantities rather than from the raw experimental peak positions.%
The second source is the 12{,}867-molecule subset of the computational OLED library for which HOMO and LUMO energies were directly available.%
Together, these two sources yield a curated fine-tuning dataset of approximately 41{,}000 molecules with consistently labelled absorption energy, oscillator strength, and HOMO--LUMO gap values.

This ordering separates the role of each dataset in the training pipeline.
ChEMBL provides an unlabeled chemical-language prior, the full computational OLED library introduces property-conditioned generation at large scale, and the curated $\sim$41{,}000-molecule dataset refines the model with higher-quality, more internally consistent electronic-structure labels.
The relationships among these datasets, including their approximate sizes, available properties, and their assignments to each training phase, are summarized in \cref{fig:oled_schematic}.

\subsection{Optical properties and conditioning}\label{sec:oled_properties}

Our conditional objectives are the vertical absorption energy $\Delta E_{S_0 \rightarrow S_1}$ and oscillator strength $f_{S_0 \rightarrow S_1}$ of the lowest singlet transition, $S_0 \rightarrow S_1$. 
We report excitation energies in eV and convert them to wavelengths through \(\lambda = hc/\Delta E\), such that lower energies correspond to red-shifted absorption and higher excitation energies to blue-shifted absorption. 
The oscillator strength is a dimensionless electronic-structure descriptor of transition intensity and measures how strongly the $S_0 \rightarrow S_1$ excitation contributes to the UV--Vis absorption spectrum.

These two quantities define the primary conditioning space because they are consistently available across the property-labelled datasets and are the same observables used for downstream TD-DFT validation.
During training and prompting, continuous values of $\Delta E_{S_0 \rightarrow S_1}$ and $f_{S_0 \rightarrow S_1}$ were converted into discrete property tokens and prepended to the SMILES sequence, as illustrated in \cref{fig:ppty_binning}.
We used five quantile bins per property, indexed from 0 to 4, such that each bin contains approximately one-fifth of the corresponding property distribution used for token construction.
Thus, token index 0 denotes the lowest-property quantile and token index 4 denotes the highest-property quantile.
For absorption, this corresponds to a progression from lower-energy, red-shifted transitions to higher-energy, blue-shifted transitions; for oscillator strength, it corresponds to increasingly intense transitions.

The HOMO--LUMO gap was included as an auxiliary electronic descriptor to regularize electronic trends during training, whereas generation and model assessment were performed primarily on the two-dimensional grid of absorption energy and oscillator strength tokens.

The numerical bin boundaries used for tokenizations are reported in \cref{tab:property_bins}.
Bins are left-closed and right-open, except for the final bin, which includes the upper boundary.
Because the bins were obtained from quantiles rather than fixed-width intervals, their widths are not uniform; this is particularly important for oscillator strength, whose distribution is strongly skewed toward small values.

\begin{table}[t]
\centering
\caption{
Quantile thresholds used for property-token assignment.
Each property was discretized into five bins using the 20th, 40th, 60th, and 80th percentiles of its distribution.
The lowest and highest bins collect values below and above the outer reported thresholds, respectively.
}
\label{tab:property_bins}
\begin{tabular}{llll}
\hline
Token index & $\Delta E_{S_0 \rightarrow S_1}$ / eV & $f_{S_0 \rightarrow S_1}$ & HOMO--LUMO gap / eV \\
\hline
0 & $< 2.7575$ & $< 5.5232\times10^{-4}$ & $< 6.2088$ \\
1 & $[2.7575,\,3.0675)$ & $[5.5232\times10^{-4},\,4.8000\times10^{-3})$ & $[6.2088,\,6.5654)$ \\
2 & $[3.0675,\,3.3489)$ & $[4.8000\times10^{-3},\,3.0500\times10^{-2})$ & $[6.5654,\,6.9263)$ \\
3 & $[3.3489,\,3.6710)$ & $[3.0500\times10^{-2},\,3.2160\times10^{-1})$ & $[6.9263,\,7.3542)$ \\
4 & $\geq 3.6710$ & $\geq 3.2160\times10^{-1}$ & $\geq 7.3542$ \\
\hline
\end{tabular}
\end{table}

Note~\ref{si:model_details}.

\subsection{Generation, filtering, and screening-set construction}

For the generation step, we sampled molecules independently for each absorption-energy/oscillator-strength conditioning pair using stochastic decoding with \ac{dcm} as the solvent condition. Unless otherwise stated, generation used \texttt{max\_new\_tokens = 150}, \texttt{num\_return\_sequences = 250}, \texttt{num\_beams = 1}, \texttt{temperature = 1.0}, and \texttt{do\_sample = true}. Generated \acp{smiles} were canonicalized, and invalid and duplicate strings were removed before downstream analysis. We define validity as the fraction of generated strings that yield chemically valid molecules, uniqueness as the fraction of distinct molecules in the full generated sample, and scaffold diversity as $1-\overline{T}$, where $\overline{T}$ is the mean pairwise Tanimoto similarity between Murcko scaffold fingerprints. Formal charges were computed on the valid unique set. For TD-DFT evaluation, we restricted attention to neutral molecules and randomly selected 20 candidates from each conditioning pair to construct a balanced screening subset.

\subsection{TD-DFT optical-property validation}\label{sec:optical}

We evaluated the optical properties of the generated candidate molecules using a semiempirical-to-\ac{tddft} workflow that combines conformer selection, \ac{dft} geometry refinement, and final excited-state calculations (\cref{alg:mf-pipeline}).

\begin{algorithm}
\caption{TD-DFT optical-property validation workflow}
\label{alg:mf-pipeline}
\begin{algorithmic}[1]
\Require Candidate molecules as SMILES

\ForAll{$s$ in candidates}
    \State Generate 3D conformer from SMILES
    \State Conformer search 
    \State Select lowest-energy conformer
   % \State Compute semiempirical properties $(\Delta E_{\mathrm{xtb}}, f_{\mathrm{xtb}})$

    \State DFT geometry optimization
    \State Frequency check
    \While{not a minimum}
        \State Displace geometry along imaginary mode(s)
        \State DFT geometry optimization
        \State Frequency check
    \EndWhile

    \State Compute high-fidelity properties $(\Delta E_{\mathrm{TD-DFT}}, f_{\mathrm{TD-DFT}})$ 
 %   \State Store $(s,\Delta E_{\mathrm{xtb}}, f_{\mathrm{xtb}}, \Delta E_{\mathrm{TD-DFT}}, f_{\mathrm{TD-DFT}})$
\EndFor

\State \Return TD-DFT optical properties for all candidates
\end{algorithmic}
\end{algorithm}

Starting from each \ac{smiles} string, initial three-dimensional geometries were generated using Open Babel.\cite{Yoshikawa2019Dec}%
Conformational sampling was then performed with \ac{crest}~2.12~\cite{crest} at the \ac{gfn2xtb} level,\cite{GFN2} using implicit solvation for \ac{dcm} through the \ac{rgbsa} model.\cite{srinivasan_continuum_1998}%
For each molecule, the lowest-energy conformer from the \acs{crest} ensemble was retained as the starting point for \acs{dft} refinement.%

The selected conformer was optimized with ORCA~6.1.0.\cite{neese2025software}
Geometry optimisations and harmonic frequency calculations were performed with the B97-3c composite method\cite{B97-3c} with \ac{rijcosx},\cite{RIJCOSX} the def2-mTZVP basis set,\cite{Weigend2005Aug} and the corresponding def2/J auxiliary basis.\cite{Weigend2006Feb}
\ac{cpcm} (\acs{dcm}) was applied consistently during geometry optimization, frequency analysis, and excited-state calculations.%
Following optimization, a vibrational analysis was performed to confirm a minimum on the potential-energy surface. When imaginary frequencies were found, the structure was displaced along the corresponding normal mode and re-optimized. This procedure was repeated until no meaningful imaginary frequency remained (Algorithm~\ref{alg:mf-pipeline}).
Optimized structures with zero or one low-lying ($<20$\,cm$^{-1}$) imaginary mode were accepted for the subsequent excited-state calculations.
Vertical excitation energies and oscillator strengths were then computed at the TD-DFT level using the long-range-corrected $\omega$B97X-D3 functional\cite{wB97X-D,wB97X-D3} and the def2-TZVP basis set,\cite{Weigend2005Aug} again with CPCM(DCM).
The reported absorption energy $\Delta E_{S_0 \rightarrow S_1}^{\mathrm{TD-DFT}}$ and oscillator strength $f_{S_0 \rightarrow S_1}^{\mathrm{TD-DFT}}$ correspond to the lowest singlet excitation.

\subsection*{Code and data availability}

The code used to train and sample the GPT-2 model for OLED generation is available at \url{https://github.com/aspuru-guzik-group/oled_generation_optical_preconditioning}. 
The workflow for TD-DFT evaluation is available at \url{https://github.com/aspuru-guzik-group/OLED_accelerated_screening}. 

The datasets generated and analyzed during the current study, together with the trained models, will be made publicly available on Zenodo upon publication.

%%%

% \section*{Author contributions}
% \input{includes/include-author-contributions}

\section*{Acknowledgments}
The authors thank Dr.~Changhyeok Choi, Dr.~Marcel M\"uller, and Benedik B\"adorf for helpful discussions.

M.G.L.\ acknowledges support from the Spanish Ministry of Science, Innovation and Universities through project PID2023-149150OB-I00, the predoctoral research contract PRE2021-098697, the ``Mar\'{\i}a de Maeztu'' Programme for Units of Excellence in R\&D (CEX2023-001316-M), and to the Vector Institute for Artificial Intelligence through the Vector Research Internship.

J.A.C.G.A acknowledges funding of this project by the National Sciences and Engineering Research Council of Canada (NSERC) Alliance Grant \#ALLRP587593-23 (Quantamole).

A.A.-G. thanks Anders G. Fr{\o}seth for his generous support. A.A.-G. also acknowledges the generous support of Natural Resources Canada and the Canada 150 Research Chairs program.

\acknowAC

\acknowSciNet[Niagara and Trillium supercomputers] % for Niagara - Trillium

\acknowCalcQueb % for Beluga or Narval

\clearpage

%%%

{
\small
\bibliography{references}
% \bibliographystyle{assets/plainnat}
% other options:
\bibliographystyle{unsrt}
}

%%%

\clearpage

\appendix
% \appendixtoc

%%%

% Reset counters
\setcounter{section}{0}
\setcounter{subsection}{0}
\setcounter{figure}{0}
\setcounter{table}{0}
\setcounter{equation}{0}

% Section numbering → S1, S2, ...
\renewcommand{\thesection}{S\arabic{section}}
\renewcommand{\thesubsection}{S\arabic{section}.\arabic{subsection}}

% Figures, tables, equations → S1, S2, ...
\renewcommand{\thefigure}{S\arabic{figure}}
\renewcommand{\thetable}{S\arabic{table}}
\renewcommand{\theequation}{S\arabic{equation}}

\section*{Supporting Information}
\label{app:related}

\Needspace{10\baselineskip}
\subsection{Supplementary material index}\label{si:contents}

\noindent\textbf{Contents overview.} The list below summarizes the sections and major items included in the Supporting Information.

\begin{figure}[!t]
    \centering
    \includegraphics[width=0.9\linewidth]{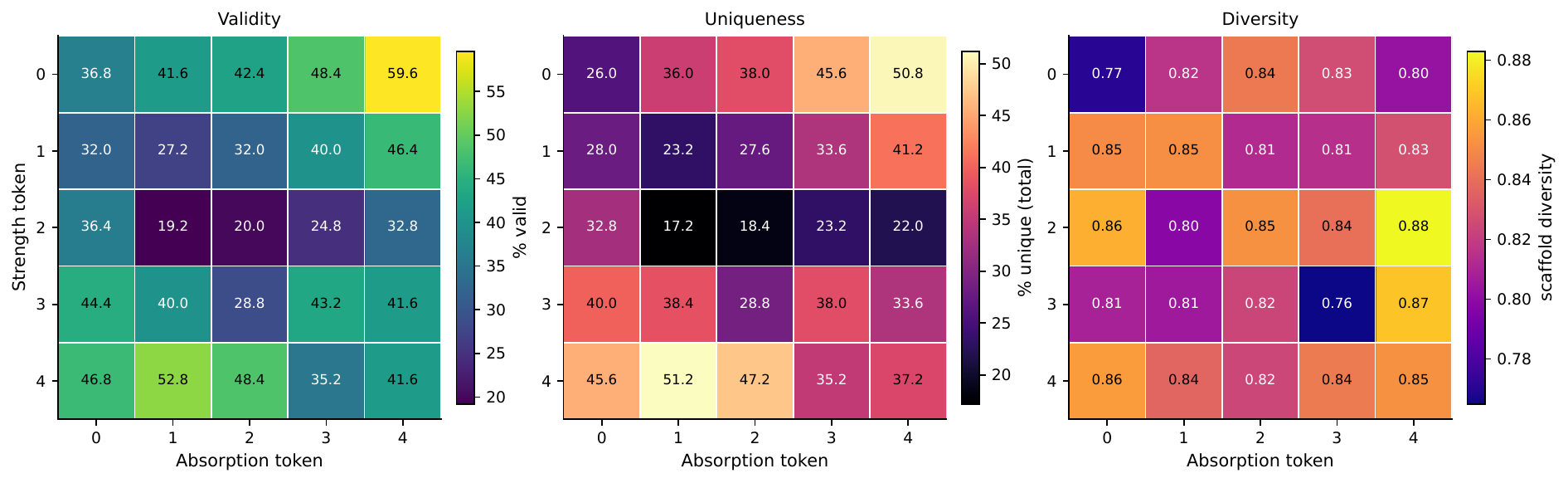}
    \caption{
    \textbf{Generation statistics across absorption-/oscillator-strength conditioning.}
    Heatmaps report the fraction of valid SMILES (left), the fraction of unique molecules in the full sample (center), and the Murcko-scaffold diversity (right), defined as $1-\overline{T}$ where $\overline{T}$ is the mean pairwise Tanimoto similarity between Murcko scaffold fingerprints, for molecules sampled at each conditioning pair.
    Validity and uniqueness vary across the target grid, indicating non-uniform calibration of conditional generation. In contrast, scaffold diversity remains consistently high, showing that the model explores chemically distinct cores across most conditioning regimes.
    }
    \label{fig:generation_metrics_token_heatmaps}
\end{figure}

\begin{figure}[!t]
    \centering
    \includegraphics[width=0.9\linewidth]{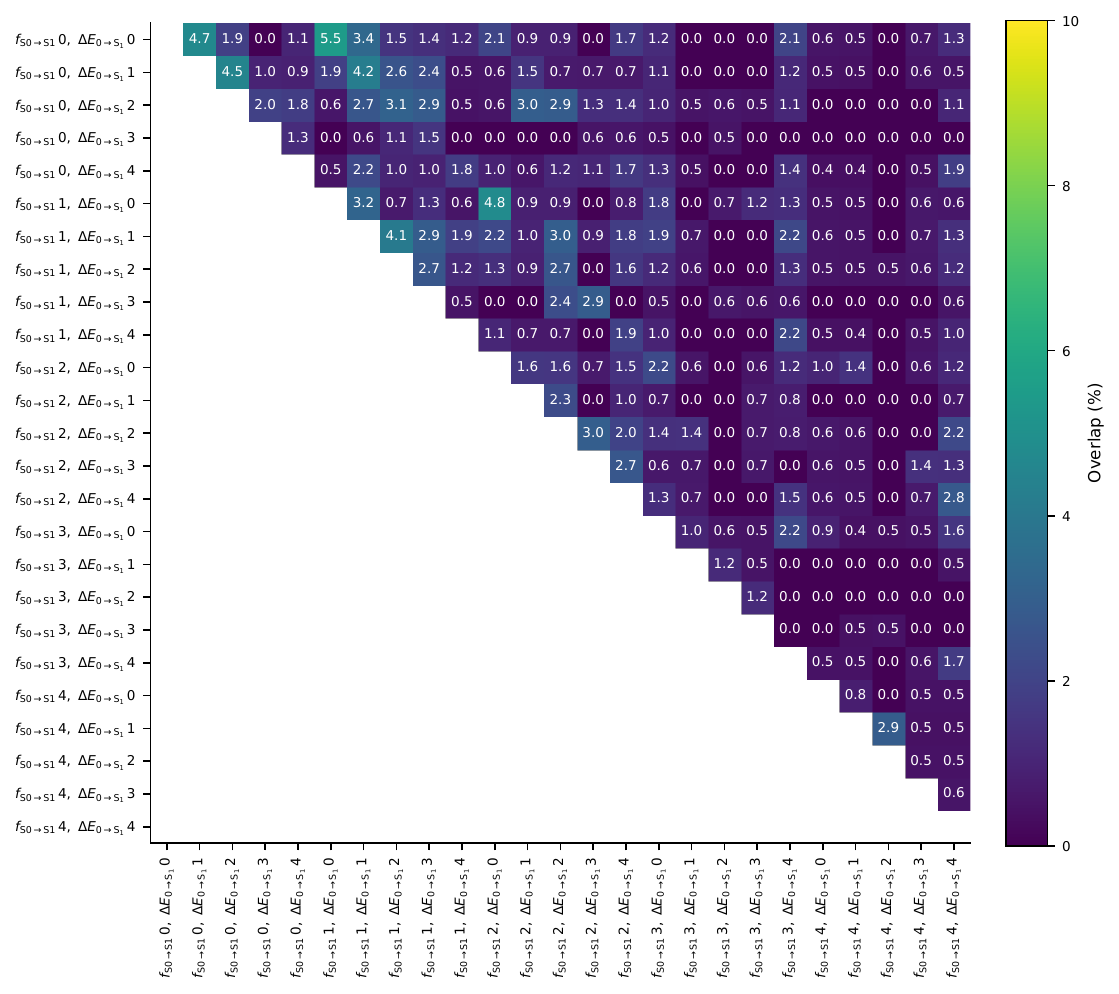}
    \caption{
    \textbf{Pairwise overlap between conditioning-specific generated sets.}
    Each cell reports the percentage overlap between the valid unique molecule sets generated under two absorption-/oscillator-strength conditioning pairs.
    Overlap is generally low, indicating that different prompts sample largely distinct regions of chemical space; modestly higher overlap is concentrated between nearby conditions, consistent with partial continuity across neighbouring target bins.
    }
    \label{fig:overlap_token_groups}
\end{figure}

\begin{figure}[!t]
    \centering
    \includegraphics[width=0.9\linewidth]{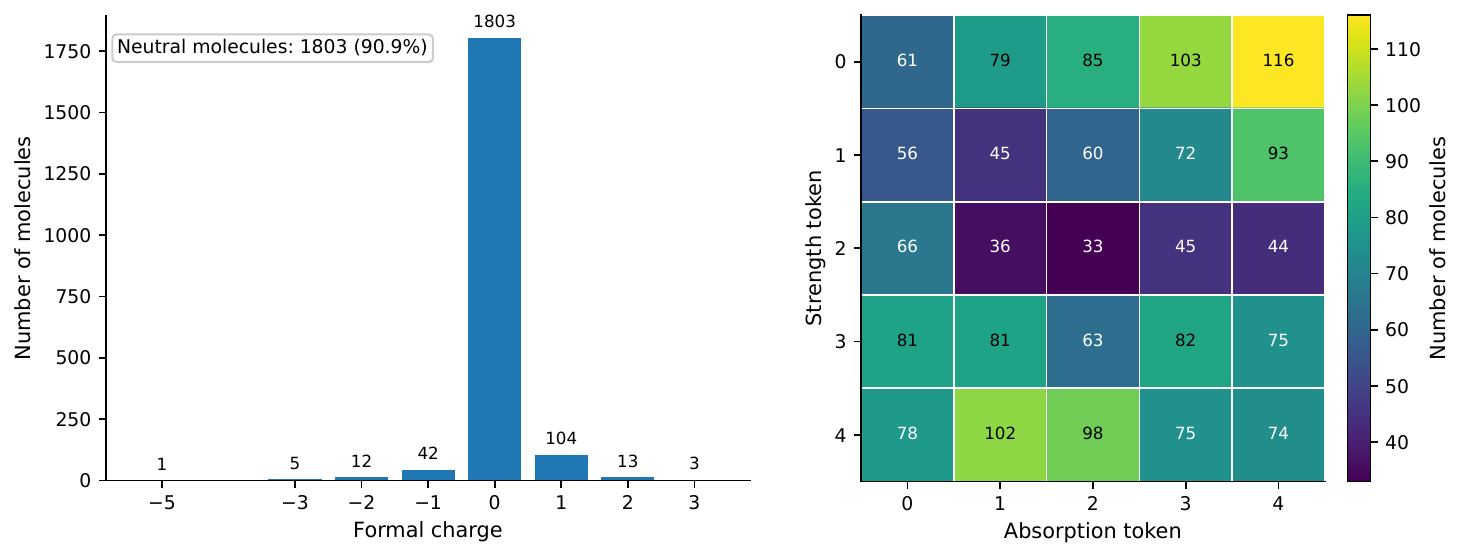}
    \caption{
    \textbf{Formal-charge composition of the generated library and number of neutral candidates by conditioning pair.}
    Left: formal-charge histogram for the valid and unique generated molecules.
    Right: number of neutral molecules retained for each absorption-/oscillator-strength conditioning pair.
    The generated set is dominated by neutral species (1803 molecules, 90.9\%), and every conditioning pair contains sufficient neutral candidates to construct a balanced validation subset of 20 molecules per pair for subsequent TD-DFT screening.
    }
    \label{fig:formal_charge_token_pair_counts}
\end{figure}

\begin{figure}[H]
    \centering
    \includegraphics[width=0.8\linewidth]{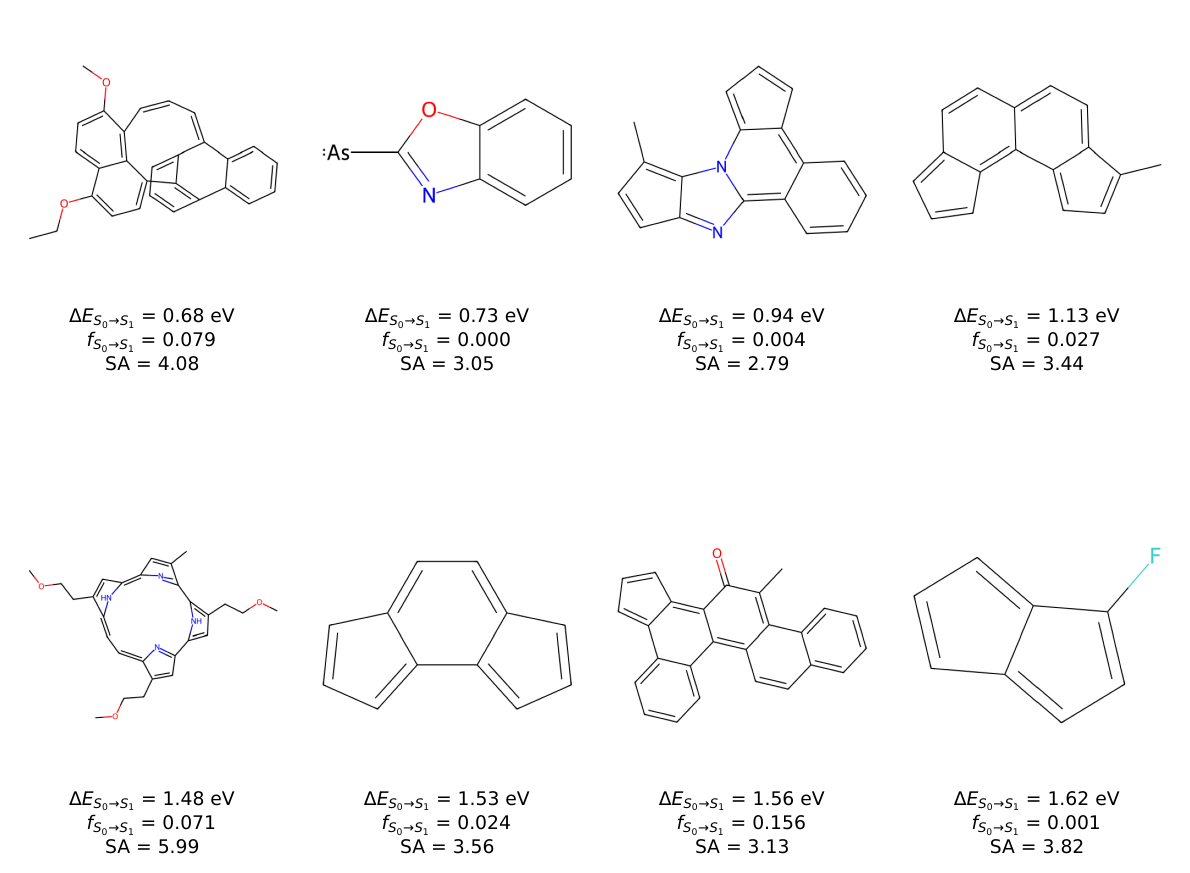}
\caption{
Infrared candidates ($\Delta E_{S_0\rightarrow S_1} < 1.63$ eV) presented in Figure~\ref{fig:training_generated_distributions_and_scatter}, ordered by increasing excitation energy.
For each molecule, the TD-DFT vertical excitation energy, oscillator strength, and RDKit synthetic accessibility (SA) score based on the Ertl--Schuffenhauer heuristic~\cite{ertl_estimation_2009} are reported; lower SA values indicate greater estimated synthetic feasibility.
Several of the more accessible candidates (SA $< 4$) feature fused or non-benzenoid $\pi$-frameworks, consistent with the extended conjugation typically required to reach the narrow-gap regime.
Among the candidates with SA $< 4$, the organoarsenic compound
($\Delta E_{S_0 \to S_1} = 0.73$~eV, SA $= 3.05$) is notable for its near-zero
oscillator strength ($f = 0.000$), suggesting that the narrow-gap regime may be
reached through dark or weakly allowed transitions in this structural class.
Synthetic accessibility of arsenic-containing $\pi$-systems would require
specialist assessment beyond the SA score heuristic.
}
    \label{fig:ir_molecules_SA}
\end{figure}

{\small
\begin{description}[leftmargin=0.34\linewidth,style=nextline,font=\normalfont]
\item[Section~\ref{si:contents}] Supplementary material index.
\item[Section~\ref{si:index_terms}] Index of terms, symbols, and parameters.
\item[Section~\ref{si:model_details}] Model architecture, training, and conditioning details.
\item[Section~\ref{si:gen_molecules_selected}] Generated molecules selected for screening.
\item[Section~\ref{si:ir_candidates}] IR candidates.
\item[Figure~\ref{fig:metrics_panels}] Generation performance across sampling temperatures.
\item[Figure~\ref{fig:diversity_panels}] Scaffold-level diversity across sampling temperatures.
\item[Figure~\ref{fig:generation_metrics_token_heatmaps}] Generation statistics across conditioning pairs.
\item[Figure~\ref{fig:overlap_token_groups}] Pairwise overlap between conditioning-specific generated sets.
\item[Figure~\ref{fig:formal_charge_token_pair_counts}] Formal-charge composition and neutral-candidate counts.
\item[Figure~\ref{fig:ir_molecules_SA}] Infrared candidates with TD-DFT properties and SA scores.
\item[Table~\ref{tab:median_properties}] Median TD-DFT absorption energies and oscillator strengths by target bin.
\end{description}
}

\Needspace{10\baselineskip}
\subsection{Index of terms, symbols, and parameters}\label{si:index_terms}

\noindent\textbf{Quick-reference index.} The list below summarizes the main acronyms, symbols, and generation parameters used throughout the manuscript and appendix.

{\small
\begin{description}[leftmargin=0.20\linewidth,style=nextline,font=\normalfont]
\item[\acs{oled}] \acl{oled}.
\item[\acs{smiles}] \acl{smiles} molecular string representation.
\item[\acs{gpt2}] Decoder-only \acl{gpt2} used here for autoregressive molecular generation.
\item[\acs{arm}] \acl{arm} objective; next-token prediction from left to right.
\item[\acs{mlm}] \acl{mlm} objective; masked-token reconstruction with bidirectional context.
\item[\acs{plm}] \acl{plm} objective; autoregressive factorization over sampled token orders.
\item[\acs{tddft}] \acl{tddft}, used here for higher-fidelity excited-state property evaluation.
\item[\acs{dft}] \acl{dft}.
\item[HOMO--LUMO gap] Energy difference between the highest occupied molecular orbital and the lowest unoccupied molecular orbital.
\item[\acs{dcm}] \acl{dcm} solvent condition used during generation and electronic-structure calculations.
\item[$s$] Candidate molecule represented as a SMILES string.
\item[$\Delta E$] Vertical $S_0\rightarrow S_1$ excitation energy (absorption energy).
\item[$f$] Oscillator strength of the targeted electronic transition.
\item[$(\Delta E_{\mathrm{xtb}}, f_{\mathrm{xtb}})$] Semiempirical optical properties computed from the low-cost screening workflow.
\item[$(\Delta E_{\mathrm{TD-DFT}}, f_{\mathrm{TD-DFT}})$] Higher-fidelity optical properties computed with TD-DFT.
\item[$c$] Property-conditioning signal used during guided generation.
\item[$\gamma$] Conditioning-strength parameter in the classifier-free-guidance-style sampling expression.
\item[$\mathbf{x}_{<t}$] All tokens preceding position $t$ in the autoregressive factorisation.
\item[\texttt{max\_new\_tokens}] Maximum number of generated tokens per sampled SMILES string.
\item[\texttt{num\_return\_sequences}] Number of candidate sequences sampled per conditioning prompt.
\item[\texttt{num\_beams}] Beam-search width; set to 1 for purely stochastic single-sequence sampling in the main generation experiments.
\item[\texttt{temperature}] Sampling-temperature parameter controlling output randomness.
\item[\texttt{do\_sample}] Flag indicating stochastic sampling rather than greedy decoding.
\item[$\overline{T}$] Mean pairwise Tanimoto similarity between Murcko scaffold fingerprints.
\item[Scaffold diversity] Defined here as $1-\overline{T}$.
\item[\acs{sa} score] RDKit \acl{sa} score based on the Ertl--Schuffenhauer heuristic; lower values indicate easier estimated synthesis.
\item[Lift] Motif-specific success-rate enrichment relative to the global baseline success rate.
\item[Success rate] Fraction of molecules in a group that satisfy both the requested absorption and oscillator-strength bins.
\end{description}
}

\Needspace{10\baselineskip}
\subsection{Model architecture, training, and conditioning details}\label{si:model_details}

\Needspace{7\baselineskip}
\subsubsection*{Architecture}
We use \ac{gpt2}~\cite{radford2019language}, a decoder-only transformer that generates sequences left to right.
The model consists of a stack of transformer blocks, each containing a masked multi-head self-attention layer followed by a position-wise feed-forward network.
Masked self-attention ensures that when predicting token $t$, the model can only attend to tokens at positions $\leq t$, enforcing the autoregressive factorization.
An input embedding layer maps each discrete token to a continuous vector, and a learned positional embedding encodes each token's position within the sequence.
The final hidden states are projected back onto the vocabulary via a linear head (with weights tied to the input embeddings) to produce next-token probabilities.
At inference time, tokens are sampled one at a time, each conditioned on the full history of previously generated tokens, naturally producing complete SMILES strings.

\Needspace{7\baselineskip}
\subsubsection*{Tokenization}
We use the \ac{smiles} tokenizer of Ref.~\citenum{schwaller_molecular_2019}, comprising tokens for atoms (\texttt{C}, \texttt{N}, \texttt{O}, \texttt{S}, etc.), bonds (\texttt{=}, \texttt{\#}), ring closures, branching parentheses, and multi-character tokens such as \texttt{Cl}, \texttt{Br}, \texttt{[C@@H]}, and \texttt{[N+]}, as well as the special tokens for sequence control (\texttt{<bos>}, \texttt{<eos>}, \texttt{<pad>}) and the property tokens.
To improve data efficiency, we apply SMILES augmentation: for each molecule, we generate two additional non-canonical SMILES via RDKit's randomized enumeration, effectively tripling the training set while teaching the model to be invariant to representation.

\Needspace{7\baselineskip}
\subsubsection*{Pretraining objectives}
We describe three objectives typically used for language pretraining: \ac{arm}, \ac{mlm}, and \ac{plm}:
\begin{align}
\max_\theta\; \sum^T_{t=0} \log p_\theta(x_t|\mathbf{x}_{<t})\tag{S1}\label{si:ARM}
\end{align}
Autoregressive language modelling~\cite{daiSemisupervisedSequenceLearning2015,radford_improving_nodate} trains the model to perform next-token prediction, factorizing the likelihood unidirectionally with each token dependent only on earlier tokens ($\mathbf{x}_{<t}$). While simple and effective, ARM does not consider bidirectional contexts.
\begin{align}
\max_\theta\; \sum^T_{t=0} \textsc{is\_masked}\cdot \log p_\theta(x_t|\mathbf{x}_{\textsc{corrupted}})\tag{S2}\label{si:MLM}
\end{align}
Masked language modelling~\cite{devlinBERTPretrainingDeep2019} randomly substitutes tokens with \texttt{[MASK]} tokens and trains the model to recover the originals. Bidirectional context is exploited, but the artificial mask tokens create a discrepancy between pretraining and finetuning, making MLM unsuitable for generative tasks.
\begin{align}
\max_\theta\; \mathbb{E}_{\mathbf{z}\sim\mathcal{Z}_T}\left[\sum^T_{t=0}\log p_\theta(x_{z_t}|\mathbf{x}_{\mathbf{z}_{<t}}) \right]\tag{S3}\label{si:PLM}
\end{align}
Permutation language modelling~\cite{yangXLNetGeneralizedAutoregressive2019} retains autoregressive factorization while sampling factorization orders so that shared parameters learn from both left and right context. In practice, PLM optimization is more complex than ARM or MLM, and the need for unrelated prefix text complicates its application to SMILES generation. We therefore focus on ARM.

\Needspace{7\baselineskip}
\subsubsection*{Language pretraining}
In the first training stage, we train the model on the ChEMBL database~\cite{gaultonChEMBLLargescaleBioactivity2012} using the autoregressive objective (\cref{si:ARM}).
Applied to molecular SMILES, this amounts to learning which chemical tokens are likely to follow a given partial structure, effectively learning the grammar of valid molecules and the distribution of common substructures.
No property information is used at this stage.

\Needspace{7\baselineskip}
\subsubsection*{Property pretraining}
In the second stage, we continue training on the 460{,}205-molecule computational OLED dataset~\cite{gomez-bombarelliDesignEfficientMolecular2016}, introducing discretized property tokens.
Each molecule's computed absorption energy, oscillator strength, and HOMO--LUMO gap are discretized into bins (4 for absorption, 4 for strength, and 5 for the gap) and prepended to the SMILES string:
\begin{center}
\texttt{<bos><strength\textit{i}><absorption\textit{j}><splitting\textit{k}>}\;$\underbrace{\texttt{c1ccc2c(c1)}\ldots}_{\text{SMILES}}$\;\texttt{<eos>}
\end{center}
Positional embeddings are set to zero at property-token positions, so the model treats them as global conditioning signals rather than sequential elements.

\Needspace{7\baselineskip}
\subsubsection*{Property-conditioned generation}
We formulate conditional generation using a classifier-free-guidance-style factorization~\cite{sanchez_stay_2023}, where sampling is biased toward a target property condition $c$:
\begin{align}
    \log p_\theta(x_t|\mathbf{x}_{<t}) + \gamma \log \frac{p_\theta(x_t|\mathbf{x}_{<t}, c)}{p_\theta(x_t|\mathbf{x}_{<t})}\tag{S4}
\end{align}
Here, $\gamma$ controls conditioning strength, allowing a tunable trade-off between unconditional chemical prior and property-directed steering. For candidate generation, we use uncertainty-guided tree-based sampling~\cite{grosseUncertaintyGuidedOptimizationLarge2024} to improve sample efficiency.

\Needspace{7\baselineskip}
\subsubsection*{Fine-tuning with multi-task learning}\label{nash}
In the final stage, we finetune on the curated dataset of approximately 41{,}500 molecules along with the computational data.
To prevent catastrophic forgetting, we train on five tasks simultaneously:
(1)~property-conditioned generation on the computational OLED set,
(2)~unconditional generation on the same set,
(3)~property-conditioned generation on the curated set,
(4)~unconditional generation on the curated set, and
(5)~unconditional generation on ChEMBL.
An imbalanced-data sampler ensures each task contributes samples at every training step, resampling from smaller datasets when they are exhausted within an epoch.

Because the tasks differ in dataset size and loss magnitude, na\"ive loss averaging leads to gradient domination by the larger tasks.
We address this with Nash Multi-Task Learning~\cite{navonMultiTaskLearningBargaining2022}, which finds Pareto-optimal task weights by solving a convex program at each step, ensuring balanced optimization across all five objectives.

\Needspace{7\baselineskip}
\subsubsection*{Training details}
We use the AdamW optimiser~\cite{adamw} with a learning rate of $1 \times 10^{-5}$ and a cosine schedule with 50{,}000 warmup steps and 250{,}000 total steps. Gradients are clipped at a maximum norm of 1.0 and accumulated over 8 steps. The model is validated every 2{,}000 steps on a held-out subset of the curated data, and the best checkpoint is selected based on validation loss.

\FloatBarrier
\Needspace{10\baselineskip}
\subsection{Optimizing decoder generation}

We compared stochastic single-sequence sampling against wide-beam decoding across the temperature range explored in the generation study. The two figures below summarize how this choice affects validity, uniqueness, and scaffold-level diversity.

\begin{center}
\begin{minipage}[t]{0.48\linewidth}
  \captionsetup{hypcap=false}
  \centering
  \includegraphics[width=\linewidth]{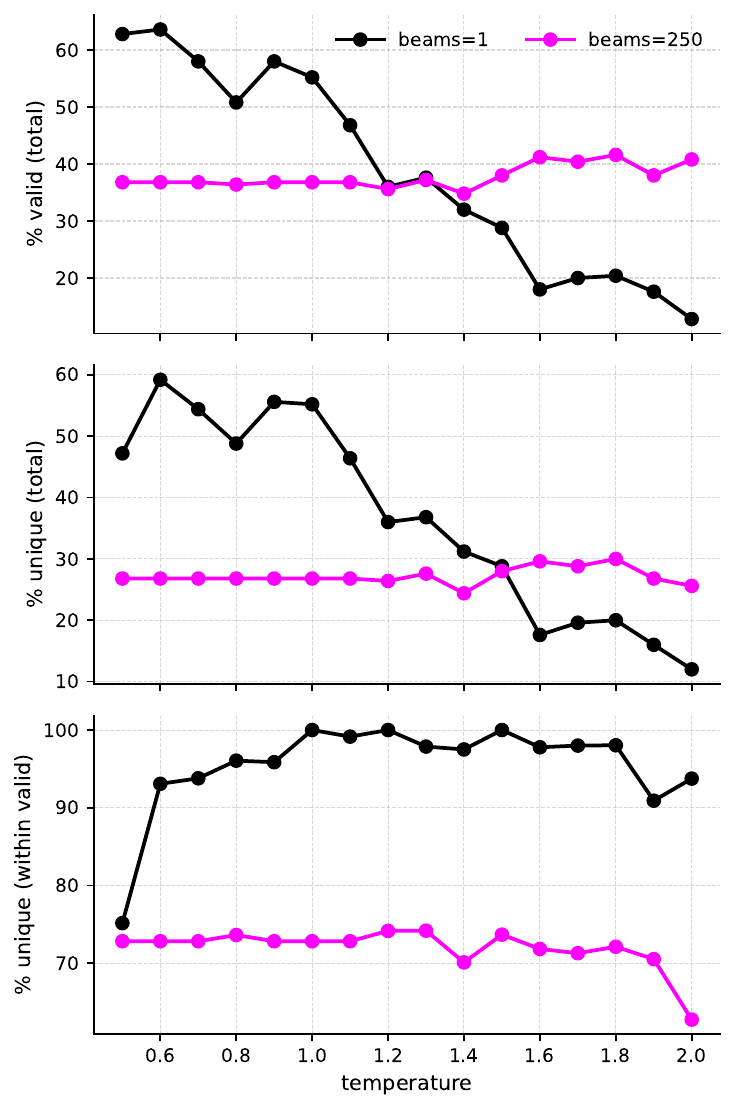}
  \captionof{figure}{\textbf{Performance of molecule generation across sampling temperatures.} Fraction of chemically valid SMILES in the full sample (a), fraction of unique molecules in the full sample (b), and fraction of unique molecules within the subset of valid SMILES (c) for \texttt{beams = 1} (black) and \texttt{beams = 250} (magenta).}
  \label{fig:metrics_panels}
\end{minipage}\hfill
\begin{minipage}[t]{0.48\linewidth}
  \captionsetup{hypcap=false}
  \centering
  \includegraphics[width=\linewidth]{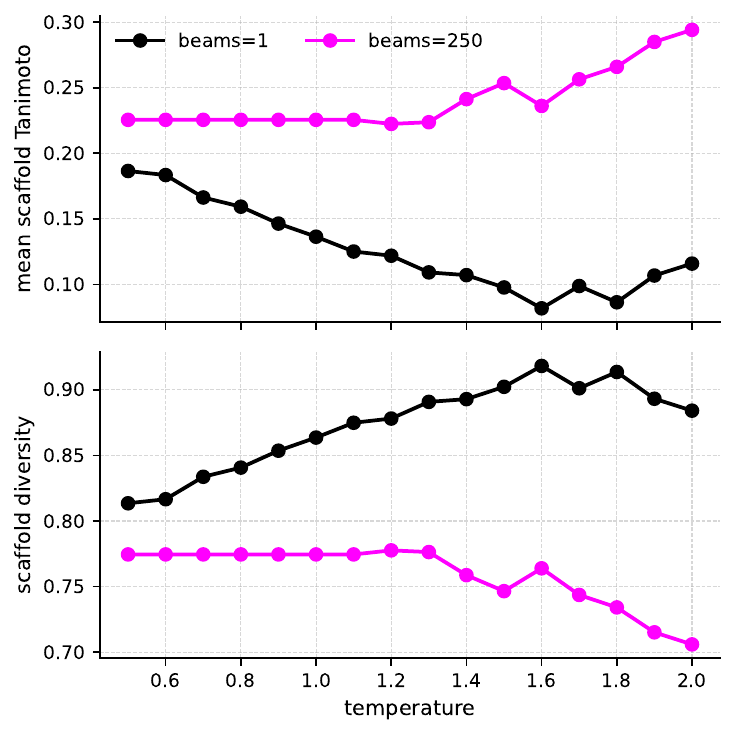}
  \captionof{figure}{\textbf{Scaffold-level diversity across sampling temperatures.} (a) Mean pairwise Tanimoto similarity between Murcko scaffold fingerprints and (b) scaffold diversity, defined as $1 - \mathrm{mean\ Tanimoto}$, for \texttt{beams = 1} (black) and \texttt{beams = 250} (magenta).}
  \label{fig:diversity_panels}
\end{minipage}
\end{center}

\FloatBarrier
\Needspace{10\baselineskip}
\subsection{Generated molecules selected for screening}\label{si:gen_molecules_selected}

To assess property-conditioned generation, we sampled molecules for each absorption-energy/oscillator-strength conditioning pair using stochastic decoding. In all cases, we used dichloromethane (DCM) as the solvent, as it is the most common solvent in the dataset. Generation was performed on CPU with \texttt{max\_new\_tokens = 150}, \texttt{num\_return\_sequences = 250}, \texttt{num\_beams = 1}, \texttt{temperature = 1.0}, and \texttt{do\_sample = true}, writing up to 250 SMILES per conditioning pair. For the quantum-chemical validation subset, we restricted attention to neutral molecules and randomly selected 20 candidates from each conditioning pair.

\begin{table}[t]
\centering
\caption{
\textbf{Median TD-DFT absorption energies and oscillator strengths obtained conditioning at the (absorption, strength) level.}
}
\label{tab:median_properties}
\begin{tabular}{cccc}
\hline
Absorption bin &  $\Delta E_{\mathrm{0\rightarrow S_1}}$
& Strength bin & Median $f_{\mathrm{S0 \rightarrow S1}}$ \\
\hline
0 & 3.469 & 0 & 0.043 \\
1 & 3.594 & 1 & 0.107 \\
2 & 3.642 & 2 & 0.069 \\
3 & 3.906 & 3 & 0.158 \\
4 & 4.601 & 4 & 0.465 \\
\hline
\end{tabular}
\end{table}

\FloatBarrier
\Needspace{10\baselineskip}
\subsection{IR candidates}\label{si:ir_candidates}

We highlight the lowest-energy infrared candidates from the TD-DFT-screened subset below, together with their excitation energies, oscillator strengths, and synthetic-accessibility estimates.

%%%

\end{document}